%% file: acl_latex.tex
\pdfoutput=1
\documentclass[11pt]{article}

\usepackage[]{acl}

\usepackage{times}
\usepackage{latexsym}

\usepackage[T1]{fontenc}
\usepackage[utf8]{inputenc}

\usepackage{microtype}

\usepackage{adjustbox}
\usepackage{amsmath}
\usepackage{amssymb}
\usepackage[scaled=0.8]{beramono}
\usepackage{booktabs}
\usepackage{multicol}
\usepackage{multirow}
\usepackage{todonotes}
\usepackage{xcolor}
\usepackage{xspace}

\newcommand{\dataset}{\textsc{SEAMuS}\xspace}
\newcommand{\famus}{\textsc{FAMuS}\xspace}
\newcommand{\mucsum}{\textsc{MUCSUM}\xspace}
\newcommand{\bos}{\texttt{$\langle$B$\rangle$}\xspace}
\newcommand{\eos}{\texttt{$\langle$E$\rangle$}\xspace}
\newcommand{\sep}{\texttt{$\langle$S$\rangle$}\xspace}

\newcommand{\zs}{\textsc{ZS}}
\newcommand{\fs}{\textsc{FS}}
\newcommand{\ft}{\textsc{FT}}
\newcommand{\textonly}{\textsc{Text Only}\xspace}
\newcommand{\eventonly}{\textsc{Event Only}\xspace}
\newcommand{\textwithschema}{\textsc{Text+Schema}\xspace}
\newcommand{\textwithevent}{\textsc{Text+Event}\xspace}
\newcommand{\ttset}[1]{$\langle$\texttt{#1}$\rangle$}

\newcommand{\conf}[3]{$[#2, #3]$}

\definecolor{best}{RGB}{16, 101, 171}
\definecolor{secondbest}{RGB}{58, 147, 195}
\definecolor{thirdbest}{RGB}{142, 196, 222}
\definecolor{fourthbest}{RGB}{246, 164, 130}
\definecolor{fifthbest}{RGB}{215, 95, 76}
\definecolor{sixthbest}{RGB}{179, 21, 41}
\title{Cross-Document Event-Keyed Summarization}

\author{William Walden\textsuperscript{\rm 1} \quad Pavlo Kuchmiichuk\textsuperscript{\rm 2} \quad Alexander Martin\textsuperscript{\rm 1} \quad Chihsheng Jin\textsuperscript{\rm 2} \\
\textbf{Angela Cao}\textsuperscript{\rm 2} \quad \textbf{Claire Sun}\textsuperscript{\rm 2} \quad\textbf{Curisia Allen}\textsuperscript{\rm 2} \quad\textbf{Aaron Steven White}\textsuperscript{\rm 2} \\
  \textsuperscript{1}Johns Hopkins University\quad \textsuperscript{2}University of Rochester \\
  \texttt{\small{\{wwalden1\}@jhu.edu}}}

\begin{document}
\maketitle
\begin{abstract}
\input{sections/00-abstract}
\end{abstract}

\section{Introduction}
\label{sec:introduction}
\input{sections/01-introduction}

\section{Background}
\label{sec:background}
\input{sections/02-background}

\section{Annotation}
\label{sec:annotation}
\input{sections/03-annotation}

% \section{Experimental Setup}
% \label{sec:setup}
% \input{sections/04-setup}

\section{Experiments}
\label{sec:experiments}
\input{sections/05-experiments}

\section{Human Evaluation}
\label{sec:analysis}
\vspace{-2mm}
\input{sections/06-analysis}

\section{Conclusion}
\label{sec:conclusion}
\input{sections/07-conclusion}

\clearpage

\section*{Limitations}
\label{sec:limitations}
\input{sections/08-limitations}

\section*{Ethics}
\label{sec:ethics}
\input{sections/09-ethics}

\clearpage
\bibliography{anthology.min, custom}
\bibliographystyle{acl_natbib}

\clearpage
\appendix
\section{Additional Examples}
\label{app:examples}
\input{appendices/additional_examples}

\section{Training and Evaluation}
\label{app:training}
\input{appendices/training}

\section{LLMs}
\label{app:llms}
\input{appendices/llms}

\section{Human Evaluation}
\label{app:human_eval}
\input{appendices/human_eval}

\section{Data \& Annotation}
\label{app:annotation}
\input{appendices/annotation}

\section{Additional Results}
\label{app:additional_results}
\input{appendices/additional_results}

\section{Use of AI Assistants}
\label{app:ai_assistants}
\input{appendices/ai_assistants}

% This is an appendix.

\end{document}

%% file: sections/00-abstract.tex
\emph{Event-keyed summarization} (EKS) requires summarizing a specific event described in a document given the document text and an event representation extracted from it. In this work, we extend EKS to the cross-document setting (CDEKS), in which summaries must synthesize information from accounts of the same event as given by \emph{multiple} sources. We introduce \dataset (\textbf{S}ummaries of \textbf{E}vents \textbf{A}cross \textbf{Mu}ltiple \textbf{S}ources), a high-quality dataset for CDEKS based on an expert reannotation of the \famus dataset for cross-document argument extraction. We present a suite of baselines on \dataset---covering both smaller, fine-tuned models, as well as zero- and few-shot prompted LLMs---along with detailed ablations and a human evaluation study, showing \dataset to be a valuable benchmark for this new task.

%% file: sections/01-introduction.tex
% Properly understanding an event often requires synthesizing accounts across multiple sources. Information gathering of this kind is an \emph{active} process, where users seek detailed information about specific aspects of an event and its participants.
Providing useful information about events requires the ability not only to extract relevant, user-specified information from documents, but also to present that information in a readable form. Drawing on this observation, \citet{gantt-etal-2024-event} recently proposed \emph{event-keyed summarization} (EKS), a task that entails summarizing a \emph{particular} event, given a document and an event representation extracted from it. EKS thus seeks to satisfy both requirements---reconciling the \emph{specific} information needs of IE end users with the more generic outputs of traditional summarization models---in order to communicate \emph{precise} information about a single event in a \emph{contextualized} and \emph{readable} form. EKS can thus be viewed as event-centric controllable summarization \citep{fan-etal-2018-controllable}, where the controlled attributes are the event and roles of interest.

\begin{figure}
    \centering
    \includegraphics[width=\linewidth]{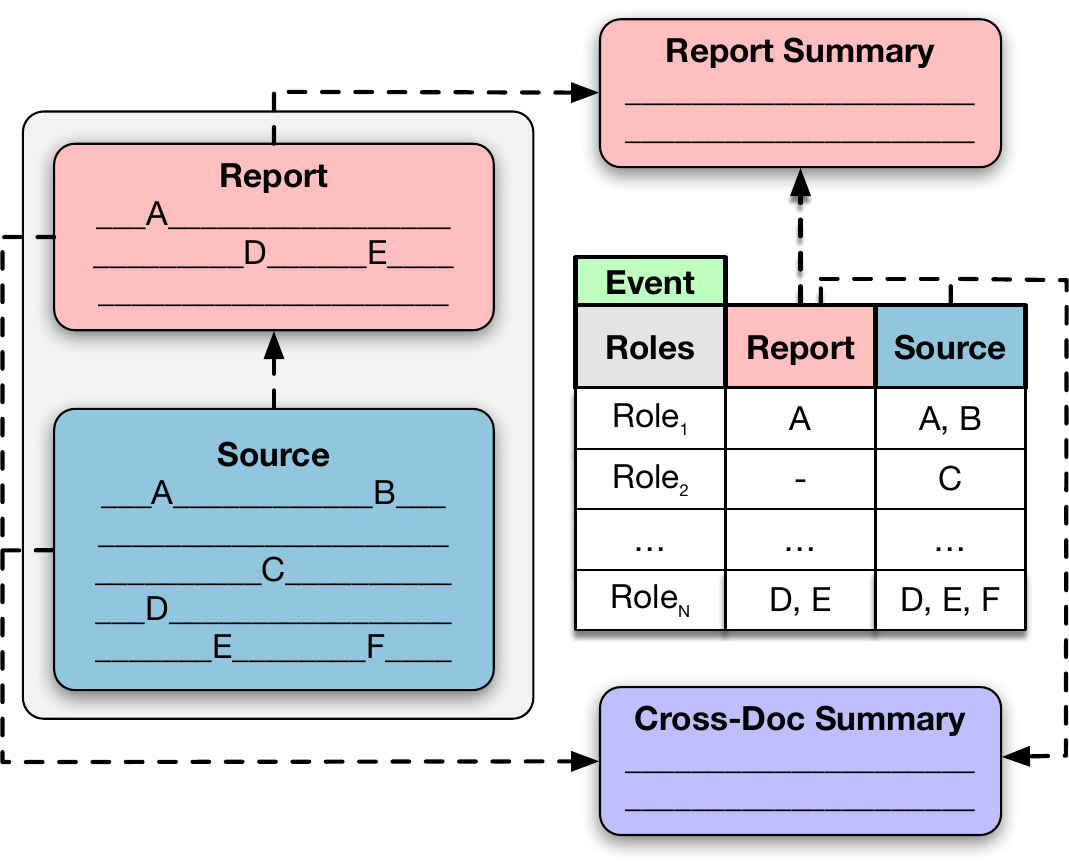}
    \caption{Schematic illustration of the \dataset report and cross-document event-keyed summarization tasks. Letters represent event arguments.\vspace{-6mm}}
    \label{fig:fig1}
\end{figure}

However, adequately understanding a particular event often requires synthesizing information across \emph{multiple} sources---evidenced in part by the rapidly growing interest in \emph{retrieval augmented generation} \citep[RAG;][]{lewis-etal-2020-retrieval}. Accordingly, this work extends EKS to the cross-document setting (CDEKS), drawing on---and enhancing---the \famus dataset for cross-document argument extraction (CDAE) to do so \citep{vashishtha-etal-2024-famus}. We summarize our contributions as follows:
\begin{enumerate}
\setlength\itemsep{-0.1em}
    \item We collect and release an expert reannotation of the \famus CDAE dataset, correcting the existing crowdsourced annotations.
    \item Based on (1), we collect and release \dataset, an expert-annotated dataset of single- and cross-document event-keyed summaries---the first ever dataset for CDEKS.\footnote{\url{https://github.com/wgantt/SEAMuS}}
    \item We present a suite of baselines on \dataset using both smaller, fine-tuned models and prompted LLMs, showing CDEKS to be challenging relative to single-document EKS.
    \item We conduct fine-grained ablations and a human evaluation, detailing CDEKS demands as a task as well as models' current capabilities.
\end{enumerate}

%% file: sections/02-background.tex
\paragraph{\famus} \citep{vashishtha-etal-2024-famus} is a dataset of short English Wikipedia passages (\emph{reports}) paired with much longer, genre-diverse English \emph{source} documents cited by those reports.\footnote{All documents are from MegaWika \citep{barham-etal-2023-megawika}.} \famus supports two tasks: (1) \emph{Source Validation} (SV), where the goal is to determine whether a candidate source document is \emph{valid for}---i.e.\ describes the same event as---an event identified in a provided report; and (2) \emph{Cross-Document Argument Extraction} (CDAE), which entails extracting arguments for an identified event from \emph{both} the report and a valid source document. \dataset builds on the \famus CDAE data, which contains 1,265 report-source document pairs (split 3:1:1 across train, dev, and test), and annotates arguments of the same target event for each document in a pair using a subset of the FrameNet ontology restricted to frames denoting events, states, or processes \citep{baker-etal-1998-berkeley}. A single, maximally ``informative'' mention is annotated for each argument, where proper names > nominal expressions > pronouns \citep[see][]{li-etal-2021-document}. In both report and source texts, arguments may be distributed across sentences.

\paragraph{Event-Centric Summarization} In introducing EKS, \citet{gantt-etal-2024-event} released \mucsum, an EKS dataset based on the classic \textsc{MUC-4} template filling dataset \citep{sundheim-1992-overview}. \mucsum contains abstractive event-keyed summaries for each event template in \textsc{MUC-4}, written so as to faithfully express the role of each template argument, plus any minimal additional context required for the summary to act as a standalone account of the event. \citeauthor{gantt-etal-2024-event} present baselines on \mucsum, and also conduct a human evaluation of model outputs, which inspires our own (\S\ref{sec:analysis}).

% Deleted footnote that followed "standalone account of the event: \footnote{E.g.\ information about the factuality of the event is essential, though it is not expressly represented in the templates used by \citet{gantt-etal-2024-event}.}

Other event-centric summarization research has focused on \emph{timeline summarization} (TLS), which constructs chronological lists of events, often with timestamps and usually based on multiple documents \citep[][\emph{i.a.}]{allan-etal-2001-temporal, chieu-lee-2004-query, li-etal-2021-timeline, rajaby-faghihi-etal-2022-crisisltlsum}. Beyond TLS, \citet{s-hussain-etal-2022-event} use extracted event-related keywords to condition single-document summarization, and integrate an event-oriented attention mechanism into BART to encourage models to cover \emph{all} events discussed. Additionally, \citet{vallurupalli-etal-2022-poque} introduce the POQue dataset, which has annotations that characterize the subevent structure of complex events in stories and the changes undergone by their participants. Among these annotations are \emph{process summaries}, which give high-level descriptions of a complex event, and \emph{change summaries}, which describe the changes experienced by a participant as a result.

% Key differences between these works and ours are (1) our focus on \emph{single} (not multiple) events \citep{s-hussain-etal-2022-event}; and (2) our use of an event ontology---meant to represent a user's information need---to control summary contents (both).\todo{ASW: I think one or two sentences expanding on these two points could be worthwhile, given the reviewer who didn't understand why we'd want to focus on a single event.}

\paragraph{Multi-Document Summarization}
CDEKS is an event-centric multi-document summarization (MDS) task. Work on MDS has pursued a variety of goals, including synthesizing reviews \citep[][\emph{i.a.}]{ganesan-etal-2010-opinosis, chu-liu-2019-meansum}, summarizing dialogues \citep[][\emph{i.a.}]{kraaij-etal-2005-ami,chen-etal-2021-dialogsum}, distilling news articles (notably, via DUC\footnote{\url{https://duc.nist.gov/}} and TAC\footnote{\url{https://tac.nist.gov/publications/index.html}}), and generating reports \citep{mayfield-etal-2024-evaluation}. \emph{Event-centric} MDS datasets include MultiNews \citep{fabbri-etal-2019-multi} and DiverseSumm \citep{huang-etal-2024-embrace}, which focus on new stories, but \dataset is most similar to Auto-hMDS \citep{zopf-2018-auto} and WCEP \citep{gholipour-ghalandari-etal-2020-large} in being built on Wikipedia articles and their sources.

CDEKS departs from all of these, however, in \emph{responding to an explicit information need}. It is thus an event-centric form of \emph{query-oriented} MDS \citep{ma-etal-2020-multi}, where a query expressing the kind of information to be summarized is provided as additional input. But whereas queries from prior work are given in natural language---e.g.\ article titles \citep{liu-lapata-2019-hierarchical} or web searches \citep{pasunuru-etal-2021-data}---ours are structured event representations, drawing on the IE tradition of leveraging event ontologies to encode information needs, and enabling extraction-to-summarization pipelines.
 
 % Second, \dataset takes some inspiration from Auto-hMDS \citep{zopf-2018-auto} and WCEP \citep{gholipour-ghalandari-etal-2020-large} in drawing on Wikipedia articles and their sources for its construction.

% Deleted footnote after ...alongside the documents.: \footnote{This contrasts with most MDS tasks, in which a summary's focus is defined implicitly by the topical similarity of the input documents (e.g.\ as determined by clustering).}

\paragraph{Our Work}
We summarize three key differences between prior work and our own. We focus on:
\begin{enumerate}
\setlength{\itemsep}{-1mm}
    \item Synthesizing information about a \emph{single} event across \emph{multiple} sources. Both multi-event (e.g.\ TLS) and single-source (e.g.\ EKS) summarization have their place, but many practical information needs depend on the rich understanding of an \emph{individual} event that is attainable only via \emph{cross-source} synthesis.
    \item Responding to a \emph{specific} event-centric information need, not \emph{generically} summarizing event-related content \citep[\emph{contra}][]{s-hussain-etal-2022-event, vallurupalli-etal-2022-poque}.
    \item Leveraging rich, structured event representations to achieve (1) and (2)---not short, unstructured queries like web searches \citep{pasunuru-etal-2021-data} or topics \citep[][\emph{i.a.}]{allan-etal-2001-temporal, rajaby-faghihi-etal-2022-crisisltlsum}.
\end{enumerate}

% First, we focus on synthesizing information about a \emph{single} event across \emph{multiple sources}: While both multi-event (e.g.\ TLS) and single-source (e.g.\ EKS) summarization have their place, many real-world information needs depend on the rich understanding of an \emph{individual} event---whether a local election or a major protest---that is attainable only via \emph{cross-source} reading. Second, we address summarization use cases where a user has a \emph{specific} information need, not a generic interest in reducing their reading burden \citep{s-hussain-etal-2022-event, vallurupalli-etal-2022-poque}. Third, we imagine this need is specified in a structured way---via extracted events---rather than with unstructured forms like web queries \citep{pasunuru-etal-2021-data} or topics \citep[][\emph{i.a.}]{allan-etal-2001-temporal, rajaby-faghihi-etal-2022-crisisltlsum}. We thus envision CDEKS as the user-facing terminus of a retrieval+event extraction pipeline.

% use an event ontology to represent this need and to drive summarization, whereas other work either has no concept of an information need \citep{s-hussain-etal-2022-event, vallurupalli-etal-2024-saga} or else relies on unstructured representations like topics \citep[][\emph{i.a.}]{allan-etal-2001-temporal, rajaby-faghihi-etal-2022-crisisltlsum}.

% Deleted footnote: \footnote{Cf.\ \citet{schuster-etal-2024-semqa} for a related QA task.}
 % \citet{jain-etal-2024-structsum}

 \begin{figure*}
    \centering
    \includegraphics[width=\textwidth]{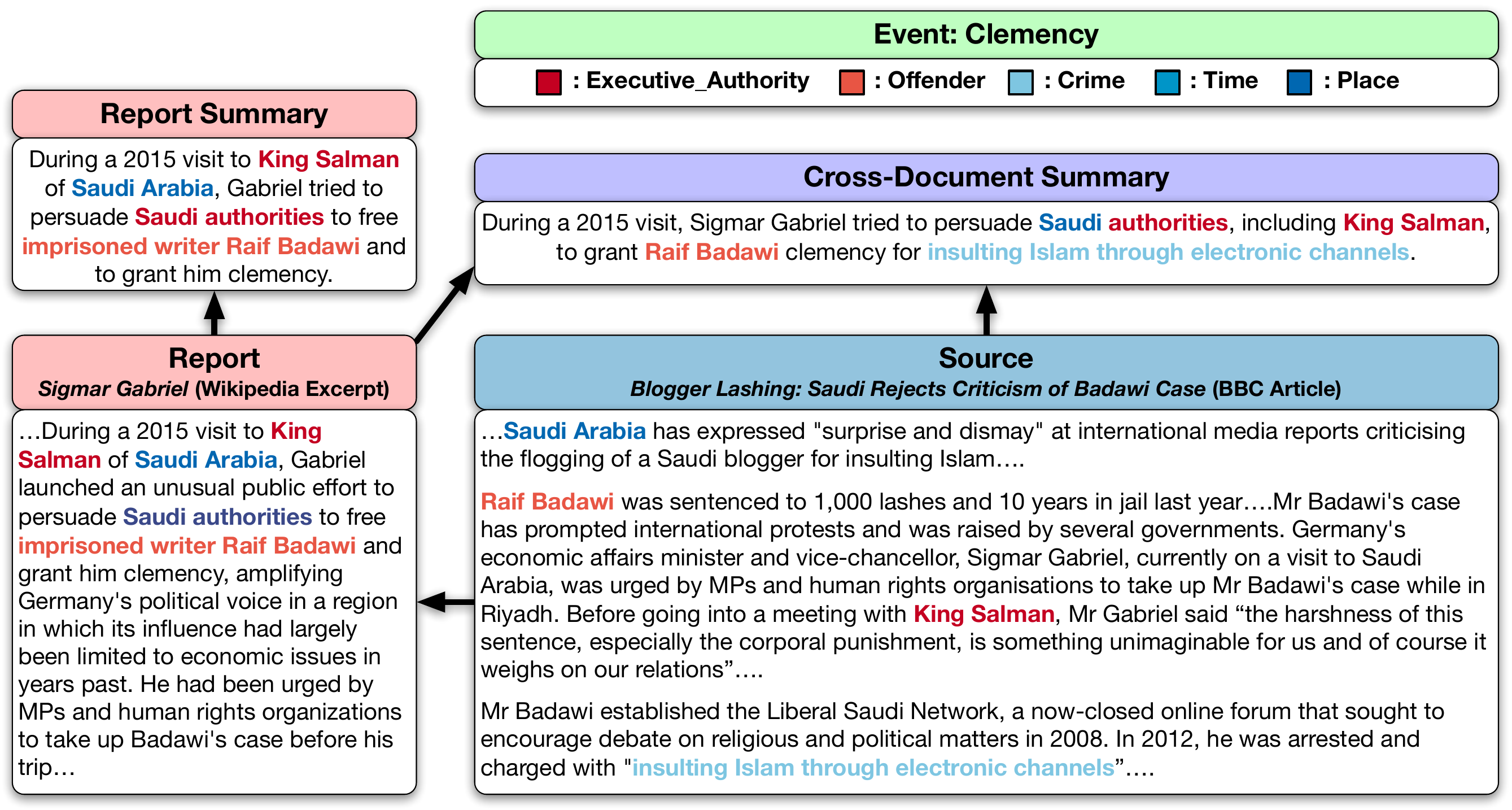}
    \caption{An example from our \dataset dataset. \textbf{Report} documents (bottom left) are Wikipedia passages that describe some event (top right) and that cite a longer (non-Wikipedia) \textbf{source} article (bottom right) as evidence, with event arguments annotated in both documents. \dataset features simple summaries of these events based on \emph{only} the report (top left) as well as enriched, cross-document summaries based on \emph{both} the report and its source, which typically contain additional information about the event (here, the \textsc{Crime}). \autoref{app:examples} has further examples.\vspace{-5mm}}
    \label{fig:fig2}
\end{figure*}

%% file: sections/03-annotation.tex
Annotation of \dataset was divided into two phases. In the first phase, abstractive \textbf{report summaries} were written for each event in \famus (see \S\ref{sec:background}) based only on its \emph{report} document, and were then annotated for event arguments (\S\ref{sec:annotation::report-summaries}). In the second phase, abstractive \textbf{cross-document summaries} were written for each event based \emph{jointly} on its report and source documents, and were then annotated for event arguments as in the first phase (\S\ref{sec:annotation::cross-doc-summaries}). In both phases, annotators were instructed to amend spurious, missing, or otherwise incorrect argument annotations in the report or source document before writing their summary. Thus, both phases involve (1) correcting existing \famus argument annotations; (2) writing a summary based on the corrected annotations; and (3) annotating arguments in the summary. The phases differ only in the documents on which the summaries are based (report only vs.\ report and source). All annotations were performed by authors of this work.\footnote{\autoref{app:annotation} has additional details and agreement results.}

\subsection{Phase 1: Report Summaries}
\label{sec:annotation::report-summaries}
Similar to the summaries in \mucsum (\S\ref{sec:background}), the report summaries in \dataset are concise summaries of a single event as recounted in a single document (a \famus report) that aim to faithfully represent the role of each participant and to provide the minimum additional context needed to serve as an accurate, standalone account of the event. Although the \famus report documents are already relatively short (typically, 2-3 sentences), they often discuss multiple events.\footnote{E.g.\ for reports in the \dataset train split, the MegaWika dataset \citep{barham-etal-2023-megawika}, from which the reports are taken, has an average of 21.4 FrameNet frames annotated.} Thus, the report summaries are further distilled descriptions focused on just \emph{one} event from the report.

Three authors completed the Phase 1 annotation, with each summary and its arguments singly annotated. Items from the train split were randomly and evenly divided among these three authors; items from the dev and test splits were similarly divided between two of them. All items were provided in JSON files containing the following information for each example: (1) a unique example ID, (2) the \famus report text; (3) the \famus-annotated frame, trigger, and arguments of the target event from the report; and (4) definitions of the annotated frame and roles as given in FrameNet. Annotators were provided with detailed instructions written by the first author and completed a 10-example practice task before beginning the main annotation. Consistent with \famus, both the corrected report arguments and the report summary arguments were annotated as single, maximally informative mentions (see \S\ref{sec:background}). Annotators were encouraged to use the same mentions in their summaries as were annotated in the (corrected) report arguments, but were permitted to alter them in the summary in order to preserve clarity or naturalness. Annotations were validated to ensure that (1) they were shorter than the report they summarized and (2) the number of arguments for a given role matched between each report and its summary. All initially invalid annotations were then corrected.

\begin{table}
    \adjustbox{max width=\columnwidth}{%
    \centering
    \begin{tabular}{l|rrrr}
    \toprule
         & \multicolumn{2}{c}{\textbf{Report}} & \multicolumn{2}{c}{\textbf{Cross-Doc}} \\
    \cmidrule(lr){2-3} \cmidrule(lr){4-5}
         & Train & Dev & Train & Dev \\
    \midrule
        Examples & 759 & 253 & 759 & 253 \\
        Avg.\ Words & 21.8 & 24.6 & 30.5 & 34.5 \\
        Avg.\ Sentences & \phantom{0}1.0 & \phantom{0}1.0 & \phantom{0}1.2 & \phantom{0}1.2 \\
        Avg.\ Arguments & \phantom{0}3.1 & \phantom{0}3.5 & \phantom{0}4.1 & \phantom{0}4.6 \\
    \bottomrule
    \end{tabular}}
    \caption{Summary statistics for the \dataset report and cross-document summaries. See \autoref{tab:additional-summary-statistics} for more. \vspace{-5mm}}
    \label{tab:summary_stats}
\end{table}

\subsection{Phase 2: Cross-Document Summaries}
\label{sec:annotation::cross-doc-summaries}
The \emph{cross-document summaries} are intended as enriched versions of the report summaries, synthesizing details about the target event from both the report and the target event's source document.

Five of the authors completed the Phase 2 annotation, with all summaries and arguments singly annotated as in Phase 1. Items from all three splits were randomly and evenly distributed to the five annotators. Given the complexity of the Phase 2 task, annotation was performed in two parts, using adapted versions of \citeauthor{vashishtha-etal-2024-famus}'s (\citeyear{vashishtha-etal-2024-famus}) interface for \famus CDAE annotation.\footnote{Interface source code was obtained from \citeauthor{vashishtha-etal-2024-famus} Screenshots are shown in \autoref{app:annotation}.}

In Part A, annotators corrected \famus argument annotations in the source documents and then wrote the cross-document summary based \emph{jointly} on the report and source texts and their corrected arguments. Annotators were encouraged to use the most informative mention of an argument across \emph{both} the report and source documents, but again were allowed to make alterations for clarity.

In Part B, annotators annotated arguments in the summaries from Part A. As in Phase 1, all annotators were provided with detailed instructions and completed a 10-example practice annotation before doing the main task. Summary argument annotations were again validated for length and to ensure that they featured as many arguments for a given role as the maximum number annotated for that role between the report and source texts.

Summary statistics for both the report and cross-document summaries can be found in \autoref{tab:summary_stats} and an example is shown in \autoref{fig:fig2}. Both types of summary average roughly a sentence in length, though cross-document summaries tend to be longer and to have more arguments---consistent with the richer information they provide.

%% file: sections/05-experiments.tex
\begin{table*}[ht]
\centering
\adjustbox{max width=\textwidth}{
\begin{tabular}{ll|lllllll|lllllll}
\toprule
& & \multicolumn{7}{c}{\textbf{Report}} & \multicolumn{7}{c}{\textbf{Cross-Document}} \\
\cmidrule(lr){1-9} \cmidrule(lr){10-16}
\textbf{Model} & \textbf{S} & $\textbf{R}_1$ & $\textbf{R}_2$ & $\textbf{R}_L$ & \textbf{BS} & \textbf{CR} & \textbf{A} & \textbf{F} & $\textbf{R}_1$ & $\textbf{R}_2$ & $\textbf{R}_L$ & \textbf{BS} & \textbf{CR} & \textbf{A} & \textbf{F}\\
\midrule
\textsc{RB} & - & $56.2$ & $46.1$ & $48.4$ & $91.6$ & $52.6$ & $99.1$ & $98.7$ & $48.5$ & $33.3$ & $39.3$ & $89.6$ & $31.0$ & $99.3$ & $93.1$\\
% \midrule
\textsc{GPT-4o M} & \zs & $62.2$ & $42.3$ & $51.3$ & $93.2$ & $58.5$ & $86.0$ & $75.8$ & $51.8$ & $29.9$ & $39.0$ & $91.3$ & $39.0$ & $81.5$ & $88.9$\\
& \fs & $72.0$ & $55.4$ & $61.0$ & $94.3$ & $66.8$ & $94.1$ & $83.3$ & $57.5$ & $36.9$ & $45.7$ & $92.1$ & $39.8$ & $88.5$ & $89.8$\\
% \midrule
\textsc{GPT-4o} & \zs & $64.0$ & $45.2$ & $53.0$ & $93.2$ & $61.4^\ast$ & $83.9$ & $74.8$ & $58.0^\ast$ & $36.4$ & $45.8$ & $92.2^\ast$ & $41.3^\ast$ & $86.6$ & $88.4$\\
& \fs & $72.5^\dag$ & $56.6^\dag$ & $62.3^\dag$ & $94.4$ & $69.6^\dag$ & $94.7$ & 81.6 & $61.2^\dag$ & $40.7^\dag$ & $49.4^\dag$ & $\mathbf{92.7}^\dag$ & $42.7^\dag$ & $90.6$ & $88.5$\\
% \midrule
\textsc{Claude H} & \zs & $64.8$ & $46.2$ & $54.7$ & $93.4$ & $58.8$ & $84.9$ & $77.6$ & $57.7$ & $36.9^\ast$ & $46.5$ & $92.1$ & $36.2$ & $90.4$ & $91.4$\\
& \fs & $71.7$ & $55.9$ & $61.1$ & $94.3$ & $63.2$ & $94.8$ & $82.5$ & $59.4$ & $39.5$ & $48.6$ & $92.1$ & $37.2$ & $91.0$ & $90.5^\dag$\\
% \midrule
\textsc{Claude S} & \zs & $67.4^\ast$ & $48.1^\ast$ & $56.5^\ast$ & $93.8^\ast$ & $61.1$ & $93.0^\ast$ & $80.6^\ast$ & $56.7$ & $34.8$ & $45.3$ & $91.9$ & $35.2$ & $93.4^\ast$ & $\mathbf{91.7}^\ast$\\
& \fs & $72.2$ & $54.6$ & $61.3$ & $94.5^\dag$ & $65.7$ & $95.9^\dag$ & $83.9^\dag$ & $57.9$ & $38.1$ & $47.4$ & $92.1$ & $37.3$ & $\mathbf{95.1}^\dag$ & $90.4$\\
% \midrule
\textsc{BART} & FT & $74.5$ & $61.7$ & $66.4$ & $94.6$ & $69.9$ & $91.6$ & $79.3$ & $63.8$ & $45.5$ & $53.0$ & $92.6$ & $\mathbf{45.0}$ & $85.6$ & $85.3$\\
% \midrule
\textsc{PEGASUS} & FT & $75.2$ & $62.5$ & $67.0$ & $94.7$ & $70.0$ & $96.1$ & $82.2$ & $63.7$ & $46.2$ & $\mathbf{53.2}$ & $92.5$ & $43.7$ & $93.9$ & $90.5$\\
% \midrule
\textsc{T5} & FT & $\mathbf{76.6}$ & $\mathbf{64.4}$ & $\mathbf{68.9}$ & $\mathbf{95.0}$ & $\mathbf{74.2}$ & $\mathbf{98.2}$ & $\mathbf{85.0}$ & $\mathbf{64.1}$ & $\mathbf{46.4}$ & $52.8$ & $92.6$ & $44.7$ & $92.5$ & $90.2$\\
\bottomrule
\end{tabular}}
\caption{\textbf{Report} and \textbf{Cross-Document} summarization results on \dataset. Best overall results are \textbf{bolded}; $^\ast$ and $^\dag$ denote best zero- and few-shot results, respectively. \textbf{S}=setting; \textsc{RB}=report baseline; \textsc{ZS}=zero-shot; \textsc{FS}=few-shot; \textsc{FT}=fine-tuned. See \S\ref{sec:experiments::overview} for an explanation of metrics; higher is better for all. See Tables \ref{tab:report-cis} and \ref{tab:cross-doc-cis} for 95\% CIs. Best \textbf{A} and \textbf{F} results exclude \textsc{RB}, for reasons explained in \autoref{app:additional_results}.\vspace{-7mm}}
\label{tab:merged-results}
\end{table*}

\subsection{Overview}
\label{sec:experiments::overview}
\paragraph{Tasks} We present experiments on both the report (\S\ref{sec:experiments::report-summarization}) and cross-document (\S\ref{sec:experiments::cross-doc-summarization}) summarization tasks. In the report task (single-document EKS), both the report and its annotated event are provided as input. The cross-document task (CDEKS) is analogous, but also includes the corresponding source document and its event annotation as input. Next, in \S\ref{sec:experiments::input-ablations}, we briefly discuss some ablations on the input inspired by similar ones from \citet{gantt-etal-2024-event}, with full results in \autoref{app:additional_results}. Finally, \S\ref{sec:experiments::extraction-quality} evaluates the impact of degraded argument extractions on summary quality.

\paragraph{Models} We benchmark \dataset using models of two types. First, we consider several classic pretrained encoder-decoder models widely used for summarization: BART \citep{lewis-etal-2020-bart}, PEGASUS \citep{zhang-etal-2020-pegasus}, and T5 \citep{raffel-etal-2020-exploring}, fine-tuning the large versions of all three on the \dataset training data. Second, we consider some of the latest proprietary LLMs, evaluated in both the zero- and few-shot settings: GPT-4o\footnote{\url{https://openai.com/index/hello-gpt-4o/}}, GPT-4o Mini (\textsc{GPT-4o M} in \autoref{tab:merged-results})\footnote{\url{https://openai.com/index/gpt-4o-mini-advancing-cost-efficient-intelligence/}}, Claude 3 Haiku (\textsc{Claude H})\footnote{\url{https://www.anthropic.com/news/claude-3-haiku}}, and Claude 3.5 Sonnet (\textsc{Claude S})\footnote{\url{https://www.anthropic.com/news/claude-3-5-sonnet}}. For the few-shot examples, we use the three examples from the train split whose frame matches that of the target example. Finally, we also give results for a \emph{report baseline} (\textsc{RB}) that treats the report text itself as the predicted summary.

\paragraph{Metrics} We report several standard summarization metrics, including ROUGE-1 ($\textbf{R}_1$), ROUGE-2 ($\textbf{R}_2$), and ROUGE-LCS $\text{F}_1$ scores \citep[$\textbf{R}_L$;][]{lin-2004-rouge}, as well as BERTScore $\text{F}_1$ \citep[\textbf{BS};][]{zhang-etal-2019-bertscore}.

Given EKS's focus on producing summaries that recover \emph{specific} pieces of information---as represented by an event's roles---we report several other metrics that evaluate this. First, we report CEAF-REE $\text{F}_1$ \citep[\textbf{CR};][]{du-etal-2021-grit}, a form of argument $\text{F}_1$ that allows us to compare arguments extracted from a \emph{predicted} summary against those in a \emph{reference} summary, aligning arguments based on exact match.\footnote{\autoref{app:additional_results} reports a soft match variant of this metric.} Following \citet{gantt-etal-2024-event}, we train the event extraction model of \citet{xia-etal-2021-lome}\footnote{\url{https://hub.docker.com/r/hltcoe/lome}} on \dataset and use it to extract arguments from the predicted summaries, constraining extraction to arguments that fill roles of the target event only.

The summaries in \dataset also make \emph{claims} about these arguments that reflect their role in the target event. To evaluate these claims' fidelity to the text, we report AlignScore \citep[\textbf{A};][]{zha-etal-2023-alignscore}, a learned metric that provides a score in $[0,1]$ that indicates how well a claim (here, a summary) is supported by a given context (the report for the report task, and the concatenated report and source for the cross-document task). We also report \textsc{FActScore} \citep[\textbf{F};][]{min-etal-2023-factscore}, which uses LMs to (1) decompose a generation into a set of \emph{atomic} facts, and (2) determine the \% of these facts supported by a given knowledge source, where \textbf{F} is the average \% supported over all examples. We use as knowledge sources the contexts used for \textbf{A}.

% FActScore \citep[\textbf{F};][]{min-etal-2023-factscore}, which decomposes each summary into a set of atomic facts and reports the average \% of these that are supported by a given knowledge source (here, the concatenated report and source texts).

% and CEAF-REE \citep{du-etal-2021-grit}. \citet{gantt-etal-2024-event} additionally report a suite of NLI-based metrics, introduced by \citet{chen-eger-2023-menli}, which compute entailment probabilities---using the predicted summary as the hypothesis and either the document or the reference summary as the premise. However, a growing body of work casts doubt on the effectiveness of such metrics, owing to the large distributional shift between the overwhelmingly sentence-level NLI datasets used to train them and the document-level texts of summarization tasks \citep[][\emph{i.a.}]{laban-etal-2022-summac, zha-etal-2023-alignscore}.

\begin{table*}
    \centering
    \adjustbox{max width=0.9\textwidth}{
    \begin{tabular}{c|c|lllllll|lllllll}
    \toprule
        & & \multicolumn{7}{c|}{\textbf{Report}} & \multicolumn{7}{c}{\textbf{Cross-Document}} \\
        \cmidrule(lr){1-9} \cmidrule(lr){10-16}
        \textbf{Model} & $p$ & $\textbf{R}_1$ & $\textbf{R}_2$ & $\textbf{R}_L$ & \textbf{BS} & \textbf{CR} & \textbf{A} & \textbf{F} & $\textbf{R}_1$ & $\textbf{R}_2$ & $\textbf{R}_L$ & \textbf{BS} & \textbf{CR} & \textbf{A} & \textbf{F}\\
    \midrule
        \multirow{6}{*}{\textsc{T5}} & 
        $0.0$ & $76.6$ & $64.4$ & $68.9$ & $95.0$ & $74.2$ & $98.2$ & $85.0$ & $64.1$ & $46.4$ & $52.8$ & $92.6$ & $46.3$ & $92.5$ & $90.2$\\
        & $0.1$ & $75.6$ & $62.8$ & $67.8$ & $93.9$ & $71.4$ & $97.6$ & $84.7$ &  $62.8$ & $45.3$ & $51.8$ & $91.5$ & $47.2$ & $92.0$ & $89.9$ \\
        & $0.2$ & $74.0$ & $61.7$ & $66.2$ & $93.6$ & $69.6$ & $98.0$ & $84.6$ & $62.0$ & $44.3$ & $50.7$ & $91.4$ & $43.5$ & $89.3$ & $88.0$ \\
        & $0.3$ & $72.1$ & $60.0$ & $64.7$ & $93.3$ & $67.5$ & $98.2$ & $83.0$ & $60.0$ & $42.8$ & $49.3$ & $91.0$ & $43.3$ & $87.3$ & $89.0$ \\
        & $0.4$ & $70.3$ & $57.5$ & $62.1$ & $92.9$ & $66.4$ & $95.8$ & $83.2$ & $58.4$ & $40.9$ & $47.8$ & $90.8$ & $44.4$ & $87.4$ & $86.8$ \\
        & $0.5$ & $68.3$ & $55.2$ & $60.6$ & $92.6$ & $63.2$ & $96.3$ & $83.5$ & $56.6$ & $39.1$ & $46.3$ & $90.4$ & $43.1$ & $87.3$ & $88.0$\\
    \midrule
        \multirow{6}{*}{\textsc{Claude H (FS)}} & $0.0$ & $71.7$ & $55.9$ & $61.0$ & $94.3$ & $63.2$ & $94.8$ & $82.9$ & $57.7$ & $36.9$ & $45.7$ & $92.1$ & $36.2$ & $91.0$ & $90.5$\\
        & $0.1$ & $67.5$ & $51.5$ & $56.7$ & $93.7$ & $59.4$ & $94.8$ & $83.6$ & $57.2$ & $37.3$ & $45.4$ & $91.8$ & $37.1$ & $82.5$ & $88.9$ \\
        & $0.2$ & $65.6$ & $48.8$ & $55.1$ & $93.5$ & $55.1$ & $94.7$ & $83.2$ & $56.2$ & $37.0$ & $45.1$ & $91.7$ & $37.8$ & $79.4$ & $88.6$ \\
        & $0.3$ & $64.7$ & $47.8$ & $54.1$ & $93.3$ & $52.8$ & $94.6$ & $84.1$ & $56.0$ & $36.2$ & $44.9$ & $91.5$ & $32.7$ & $82.2$ & $89.2$ \\
        & $0.4$ & $64.1$ & $47.2$ & $54.1$ & $93.3$ & $52.2$ & $95.0$ & $83.1$ & $54.5$ & $34.3$ & $43.1$ & $91.3$ & $31.4$ & $85.0$ & $89.0$ \\
        & $0.5$ & $63.1$ & $46.8$ & $54.0$ & $93.1$ & $52.3$ & $94.7$ & $83.8$ & $54.3$ & $34.6$ & $43.3$ & $91.3$ & $33.1$ & $86.4$ & $89.2$ \\
    \midrule
        \multirow{6}{*}{\textsc{GPT-4o M (FS)}} & $0.0$ & $72.0$ & $55.4$ & $61.0$ & $94.3$ & $66.8$ & $94.1$ & $83.3$ & $57.5$ & $36.9$ & $45.7$ & $92.1$ & $39.8$ & $88.5$ & $89.8$\\
        & $0.1$ & $69.2$ & $52.8$ & $59.5$ & $94.0$ & $64.0$ & $94.5$ & $81.8$ & $58.8$ & $38.2$ & $46.2$ & $92.1$ & $42.2$ & $74.5$ & $90.6$\\
        & $0.2$ & $67.6$ & $50.8$ & $57.0$ & $93.7$ & $59.8$ & $94.2$ & $84.3$ &$56.6$ & $36.2$ & $45.1$ & $91.2$ & $39.4$ & $75.2$ & $89.8$\\
        & $0.3$ & $66.9$ & $50.1$ & $57.0$ & $93.7$ & $59.3$ & $94.9$ & $81.8$ & $56.4$ & $36.2$ & $44.5$ & $91.8$ & $37.8$ & $77.2$ & $90.2$\\
        & $0.4$ & $65.2$ & $48.1$ & $54.9$ & $93.4$ & $56.7$ & $93.8$ & $84.2$ & $54.8$ & $34.0$ & $42.8$ & $91.6$ & $36.6$ & $77.8$ & $90.6$\\
        & $0.5$ & $65.1$ & $47.5$ & $54.8$ & $93.4$ & $55.2$ & $95.4$ & $82.7$ & $54.2$ & $33.2$ & $42.6$ & $91.4$ & $34.4$ & $80.9$ & $90.8$\\
    \bottomrule
    \end{tabular}}
    \caption{Performance of three models from \autoref{tab:merged-results} when the argument annotations for each role in the report event (\textbf{Report}) or additionally in the source event (\textbf{Cross-Document}) are corrupted with probability $p$ (see \S\ref{sec:experiments::extraction-quality}).\vspace{-5mm}}
    \label{tab:corruption-merged}
\end{table*}

\subsection{Report Summarization}
\label{sec:experiments::report-summarization}
\paragraph{Setup} As input for BART, PEGASUS, and T5, we provide the full report text concatenated with a linearized representation of the annotated report event that contains the frame name, the event trigger, and the role names, each followed by a list of the arguments annotated for that role. We train each model against a standard conditional language modeling objective w.r.t.\ the gold report summaries for a maximum of 30 epochs, using a patience of 5 epochs, with dev $\textbf{R}_1$ as the stopping criterion.\footnote{Details on training and input formats are in \autoref{app:training}.} For inference, we use beam search decoding with a beam size of 5 and a max of 256 new tokens.

% $$\text{Frame: }\langle\text{frame name}\rangle\text{ Trigger: }\langle\text{trigger}\rangle\langle\text{Role 1}\rangle\langle $$

For the Claude and GPT models, our system prompt asks the model to analyze and summarize a specific event. The user prompt provides more detailed task instructions, followed by the full report text, and a description of the target event that includes (1) the frame name and definition from FrameNet; (2) the trigger; and (3) a bulleted list, where each item includes a role name, its definition, and the arguments annotated for that role. In the few-shot setting, we format the three few-shot examples (see \S\ref{sec:experiments::overview}) the same way, but with the target summary shown at the end of each. We set temperature to 0.7 and the max new tokens to 256, leaving other API defaults unchanged.\footnote{\autoref{app:llms} has further details on models and prompts.}

\paragraph{Results} are shown in the left half of \autoref{tab:merged-results}. First, we find that T5 obtains the best performance across all metrics, followed by PEGASUS and BART, with T5 exhibiting particularly strong results for \textbf{CR}, indicating its ability to accurately recover event arguments in its summaries. Second, the LLMs almost universally outperform the report baseline (\textsc{RB})---even in the zero-shot setting (\textsc{ZS}), where Claude Sonnet generally obtains the best results. Third, adding just three few-shot examples (\textsc{FS}) yields major gains over the zero-shot setting for all LLMs on all metrics. Even here, however, few-shot results still trail the best fine-tuned results (T5) by sizable margins on most metrics.

\subsection{Cross-Document Summarization}
\label{sec:experiments::cross-doc-summarization}
\paragraph{Setup} The setup for the cross-document task is similar to that of the report task, but adds the source text and its annotated event to the input alongside the report text and its event. As the source texts are full web articles, most are long (e.g.\ dev texts average almost 62 sentences and over 1,500 words). While this is no obstacle for the LLMs, the smaller models do not support contexts of this size. Thus, to enable a fair comparison across models, we apply a sentence retriever to the source, using the report text as a query to select the top $k$ most relevant sentences to use as context.\footnote{This approach can also be justified by the fact that typically only a small portion of the source concerns the event.} We consider $k \in \{3, \ldots, 10\}$ and selected the maximum value such that $\geq 95\%$ of the resulting dev set contexts would fit untruncated in the input, yielding $k=7$. We experimented with the dense retrievers \texttt{all-mpnet-base-v2} \citep[based on MPNet;][]{song-etal-2020-mpnet} and \texttt{e5-large-v2} \citep{wang-etal-2022-text}, but obtained our best results with BM25 \citep{robertson-etal-2009-probabilistic}, which we use in all experiments.\footnote{Models were evaluated on recall of annotated arguments in the retrieved contexts for the dev set for fixed $k$. At $k=7$, BM25 recovered $\sim76\%$ of annotated source arguments.}

We use the same training and inference settings from \S\ref{sec:experiments::report-summarization}; see Appendices \ref{app:training}, \ref{app:llms} for further details.

% \paragraph{Orion's Setup}
% Following \citet{hou2024clercdatasetlegalcase}, we use a llama based retriever \cite{ma2023finetuningllamamultistagetext} for sentence and paragraph retrieval. 
\paragraph{Results} are shown in the right half of \autoref{tab:merged-results} and are qualitatively similar to those for the report task, with the fine-tuned models generally showing the best overall numbers ($\textbf{R}_{1,2,L}$, \textbf{CR}) or nearly so ($\textbf{BS}$), although GPT-4o obtains the highest scores on \textbf{BS} and Claude Sonnet on \textbf{A} and \textbf{F}. Once again, nearly all models outperform the report baseline across the board (\textsc{ZS} and GPT-4o Mini excepted). Finally, we note that results on most metrics are much lower in absolute terms compared to the corresponding results from \S\ref{sec:experiments::report-summarization}, testifying to the greater difficulty of the cross-document task.

\subsection{Input Ablations}
\label{sec:experiments::input-ablations}
Following \citet{gantt-etal-2024-event}, \autoref{app:additional_results} considers ablations on the input for both tasks, in which we omit the annotated events (\textsc{Text Only}) or the texts (\textsc{Event Only}), and condition summary generation on the resulting ablated inputs. We also present a novel third ablation that omits the \emph{annotated arguments}, but leaves intact information about the target frame and roles (\textsc{Text+Schema}). Consistent with \citeauthor{gantt-etal-2024-event}'s results, we find that both the text and the full event annotations are needed to obtain the best results (see Tables \ref{tab:report-only-ablation-results} and \ref{tab:combined-ablation-results} in \autoref{app:additional_results}), indicating that the \dataset tasks are \emph{not} reducible to standard summarization (\textsc{Text Only}), structure-to-text (\textsc{Event Only}), or even a (simpler) hybrid objective (\textsc{Text+Schema}).

% \todo{ASW: I almost wonder if this section should come before the stuff on input extraction quality. Basically, we say ``consistent with Gantt et al., we actually need arguments.'' Then, we say ``how much noise in the arguments can the models tolerate?''}

% The preceding experiments conditioned summary generation on gold event annotations in order to focus model comparisons narrowly on summarization ability. In most practical settings, however, a (CD)EKS model would instead have to rely on the imperfect outputs of an event extraction model.

\subsection{Impact of Extraction Quality}
\label{sec:experiments::extraction-quality}
\paragraph{Setup} Given that models require event arguments in the input to produce the best summaries (\S\ref{sec:experiments::input-ablations}), a natural next question concerns the sensitivity of these models to noise in the arguments. In real-world scenarios, one generally will \emph{not} have access to gold arguments (as in \S\ref{sec:experiments::report-summarization}-\ref{sec:experiments::cross-doc-summarization}) and must instead rely on the outputs of an event extraction model.

To probe robustness to extraction errors in a controlled manner, we apply variable amounts of noise to the gold event annotations and evaluate model performance on the resulting inputs. Concretely, for each role $R$ of each event, we edit $R$'s arguments with probability $p$. If a role is selected for editing, we then make \emph{one} of the following edits with equal probability:
\begin{enumerate}
    \setlength\itemsep{-0.5em}
    \item \textsc{Insert}: A new (incorrect) span from the text is \emph{added} to the argument list for $R$.
    \item \textsc{Delete}: An argument span is \emph{removed} at random from the argument list for $R$.
    \item \textsc{Replace}: An argument span is \emph{replaced} at random with an incorrect span from the text.
\end{enumerate}
For the cross-document task, we apply these edits to the event annotations for both the report and the source. We sample the edits to be made uniformly and then prompt an LLM (GPT-4o) to apply them by supplying in a prompt: (1) the text (report or source), (2) the (JSON-formatted) report or source event annotations, and (3) instructions for the edits to be made, generated automatically by populating templatic statements based on the edits sampled. The LLM is free to select an appropriate \emph{new} span to be used for the \textsc{Insert} and \textsc{Replace} operations. We consider $p \in \{0.1, 0.2, 0.3, 0.4, 0.5\}$, using the same sampled edits for all models for a given $p$.

We evaluate one fine-tuned model (T5) and one model each from the Claude (Claude Haiku) and GPT (GPT-4o Mini) families. For T5, we use the same checkpoint as presented in \autoref{tab:merged-results}. For Claude Haiku and GPT-4o Mini, we use few-shot prompts similar to those used in the \textsc{FS} setting in \autoref{tab:merged-results}, but with two key changes. First, we alter the task instructions to say that the event annotations for the target example \emph{may contain errors}, and that the model must correct these errors when generating its summary by consulting the text(s). Second, we show the model how to do this by substituting noised versions of the event annotations in the few-shot examples while leaving their associated texts and summaries unchanged.
\vspace{-2mm}

\paragraph{Results} for both tasks are in \autoref{tab:corruption-merged}. For all models, we observe (near-)monotonic drops in performance for most metrics as $p$ increases. While performance drops are sizable in some cases, they are arguably less radical than we might expect, given the destructiveness of the changes at $p=0.5$, where roughly half of all roles contain extraction errors. This is especially evident in the results for Claude Haiku and GPT-4o Mini on the cross-document task, where (e.g.) $\textbf{R}_{1,2,L}$ scores decrease by only about 3 points from $p=0$ to $p=0.5$, $\textbf{BS}$ by less than 1, and $\textbf{F}$ showing no drop at all. Further, losses on $\textbf{CR}$ (the most explicit measure of extraction ability) are only $\sim$5 points for GPT-4o Mini and $\sim$3 points for Claude Haiku.

These findings are confirmed by manual inspection of model outputs, where we often see minimal degradation in summary quality (\autoref{tab:corruption-example})---suggesting an intriguing strength of this task relative to traditional event extraction: the ability to \emph{counteract extraction errors post-hoc} by using imperfect event extractions as a query to locate relevant passages in the input and then relying on those passages to avoid analogous errors in the summary.

\begin{table}
    \centering
    \adjustbox{max width=\linewidth}{
    \begin{tabular}{l|l}
    \toprule
        $p$ & \textbf{Summary} \\
    \midrule
        \textcolor{best}{\textbf{$0.0$}} & \textcolor{best}{\textbf{The gradual accumulation of partially decayed plant material}}\\
        & \textcolor{best}{\textbf{in a bog functions as a carbon sink.}} \\
        \textcolor{secondbest}{\textbf{$0.1$}} & \textcolor{secondbest}{\textbf{The gradual accumulation of decayed plant material}} \\
        & \textcolor{secondbest}{\textbf{in a bog functions as a carbon sink.}} \\
        \textcolor{thirdbest}{\textbf{$0.2$}} & \textcolor{thirdbest}{\textbf{The gradual accumulation of decayed plant material}} \\
        & \textcolor{thirdbest}{\textbf{in a bog acts as a carbon sink.}} \\
        \textcolor{fourthbest}{\textbf{$0.3$}} & \textcolor{fourthbest}{\textbf{The gradual accumulation of decayed plant material}} \\
        & \textcolor{fourthbest}{\textbf{in a bog functions as a carbon sink.}} \\
        \textcolor{fifthbest}{\textbf{$0.4$}} & \textcolor{fifthbest}{\textbf{The gradual accumulation of decayed plant material,}}\\
        & \textcolor{fifthbest}{\textbf{including peat, in bogs functions as a carbon sink.}}\\
        \textcolor{sixthbest}{\textbf{$0.5$}} & \textcolor{sixthbest}{\textbf{The gradual accumulation of decayed plant material}}\\
        & \textcolor{sixthbest}{\textbf{in a bog functions as a carbon sink.}}\\
    \bottomrule
    \end{tabular}}
    \caption{Example outputs from GPT-4o Mini on the cross-document task as role annotations are corrupted with probability $p$. In many cases (as here), we find minimal degradation in quality from $p=0$ to $p=0.5$.\vspace{-6mm}}
    \label{tab:corruption-example}
\end{table}

%% file: sections/06-analysis.tex
\begin{figure*}
    \centering
    \includegraphics[width=0.48\textwidth]{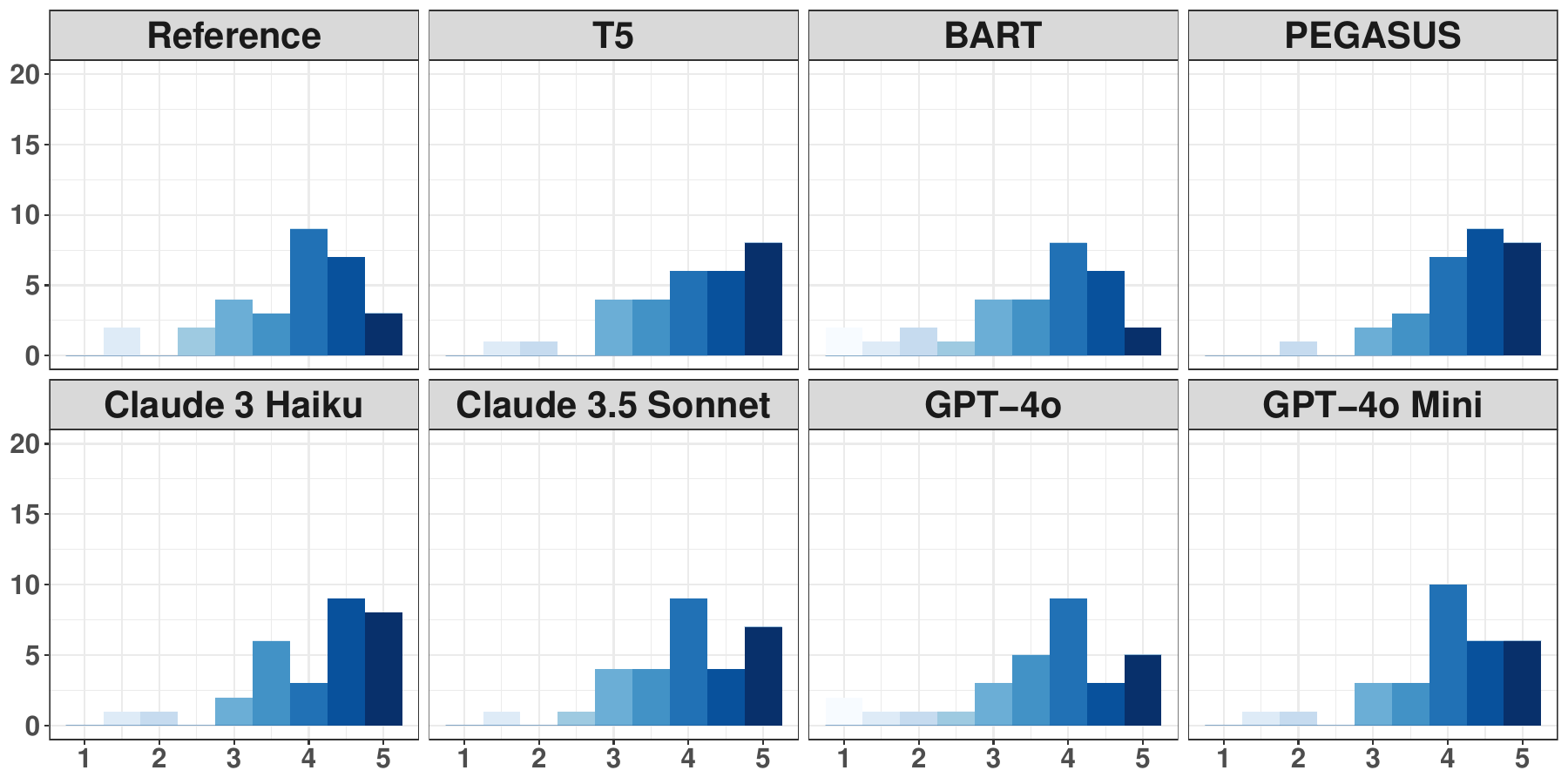}
    \includegraphics[width=0.48\textwidth]{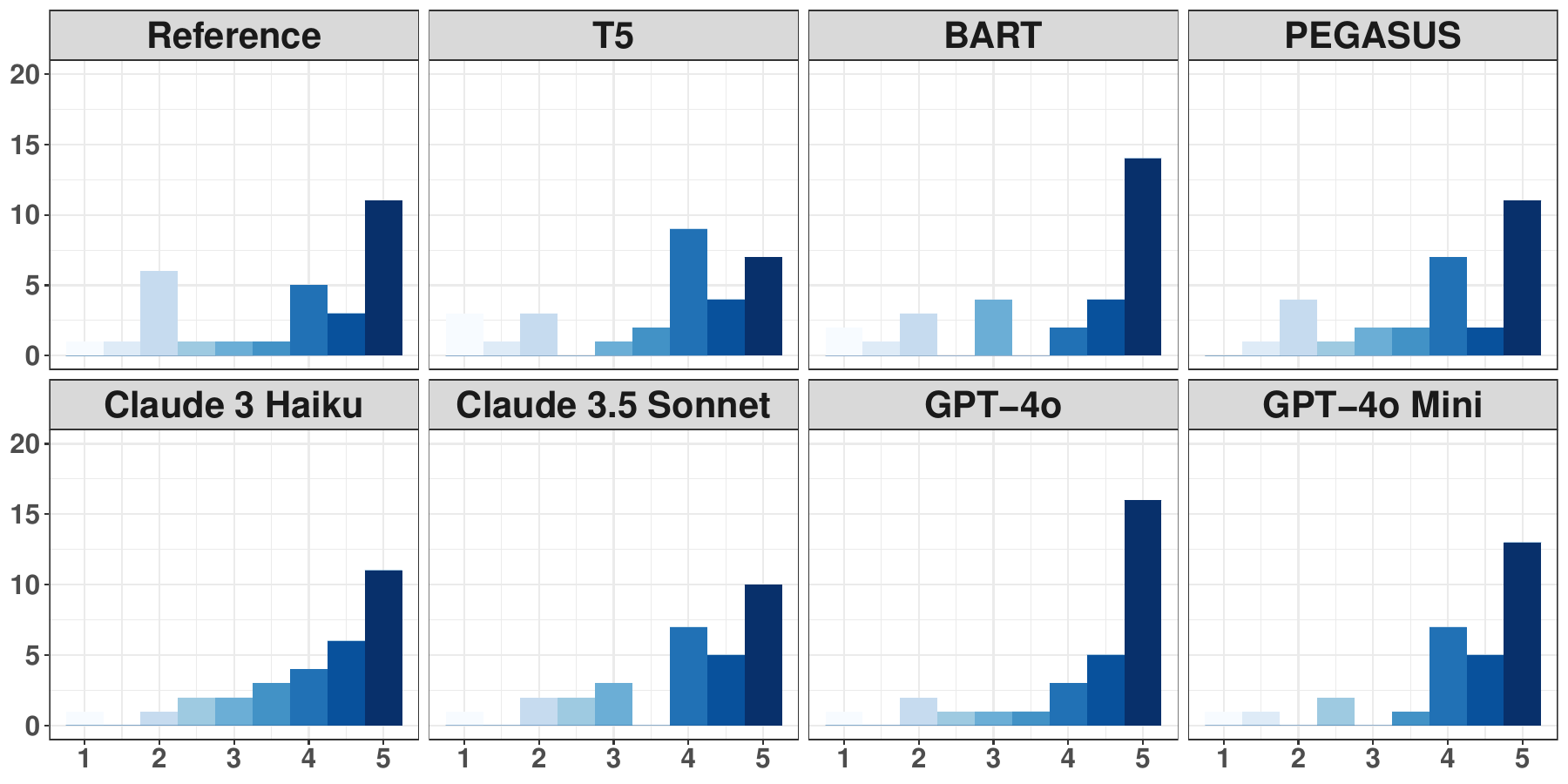}
    \includegraphics[width=0.48\textwidth]{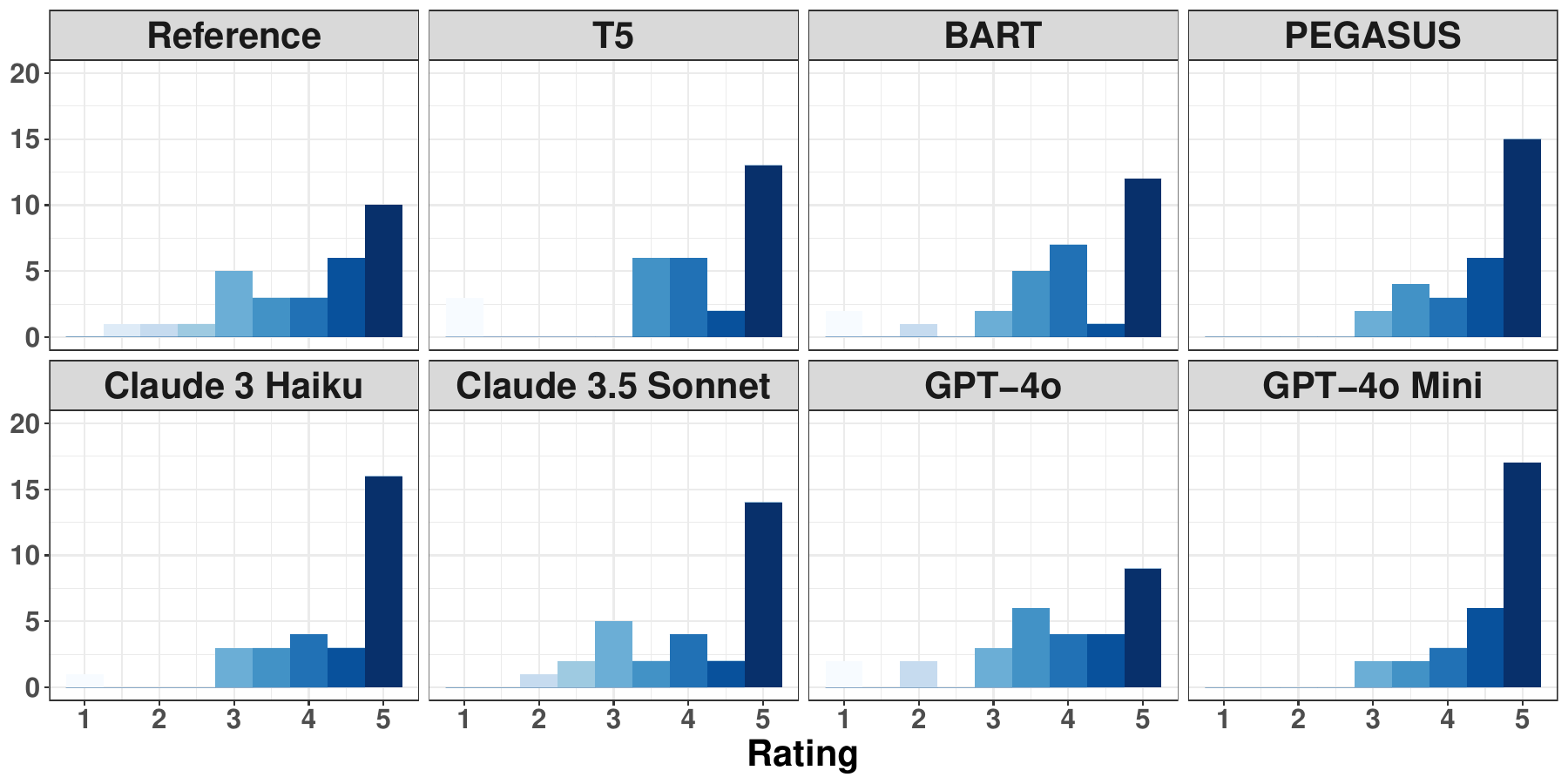}
    \includegraphics[width=0.48\textwidth]{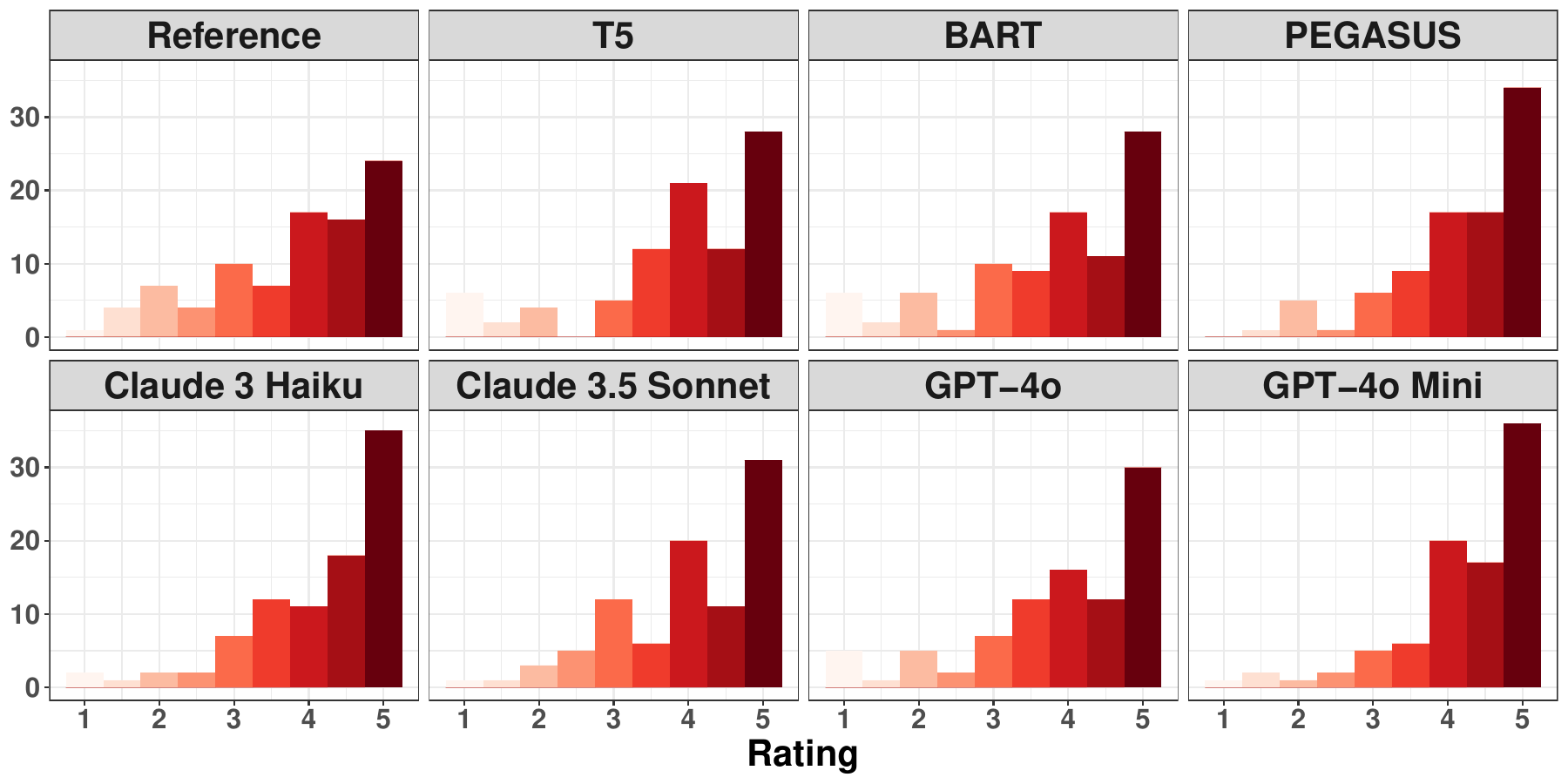}
    \vspace{-3mm}
    \caption{Histograms of summary quality scores (1-5, higher is better) from our human evaluation (\S\ref{sec:analysis}). The bottom right plot (red) aggregates scores across all three raters; each of the other plots (blue) shows a single rater's scores.\vspace{-5mm}}
    \label{fig:score-hist}
\end{figure*}

\paragraph{Setup} Lastly, we conduct a human evaluation of the reference and model-generated summaries. We focus our evaluation on the cross-document task, comparing the summaries generated by models presented in \autoref{tab:merged-results} (excluding \textsc{RB}). For the GPT and Claude models, we use the \textsc{FS} (few-shot) summaries only, owing to their superiority over the \textsc{ZS} results. We randomly sampled 30 test set examples and presented the 7 model-generated summaries for these examples, along with the references, to 3 human raters---all English-speaking NLP researchers who did not participate in other parts of this work. Each rater provided a single quality score for each summary based on the ordered list of attributes used by \citet{gantt-etal-2024-event}: \emph{factuality}, \emph{adequacy}, \emph{coherence}, \emph{relevancy}, and \emph{fluency}. Scores were given from 1 (low) to 5 (high), with half points allowed. Each rater thus provided $30 \times (7 + 1) \times 1 = 240$ judgments. Raters were not shown which model produced which summary, and summary presentation order was randomized.\footnote{See \autoref{app:human_eval} for further details.}\vspace{-1mm}

\paragraph{Results} Four sets of histograms of scores for each model (and the reference) are shown in \autoref{fig:score-hist}. The bottom right set (red) shows scores aggregated across annotators, while the other three (blue) each show scores of a single rater. For all raters, scores are consistently high across models and the reference, with modes of $\geq 4$ for each. Comparing preferences across raters, however, we see significant variability: GPT-4o achieved the highest average score for one rater (top right, 4.28); GPT-4o Mini for the second (bottom left, 4.57); and PEGASUS for the third (top left, 4.23).

% To obtain more fine-grained comparisons, we also follow \citeauthor{gantt-etal-2024-event} in conducting paired Wilcoxon rank sum tests for each pair of models, evaluating the difference between an annotator's ratings on the summaries generated by each model in a pair.

Looking at intra-rater distributions, however, it's unclear how robust these preferences are. Using Wilcoxon rank-sum tests to evaluate pairwise differences in each rater's scores for a given pair of models, we find that some of these preferences are reliable at $\alpha = .05$ (e.g. GPT-4o $>$ T5 with $p=.016$ for the first rater), but none holds up when applying the Bonferroni correction for multiple comparisons. We take these results to indicate that our baselines are fairly effective at producing good summaries, and that while they may somewhat differentiate themselves on individual metrics\footnote{See our discussion of argument recovery in \autoref{app:additional_results}.}, the best models on a more holistic picture may come down to user preference, and there may not be \emph{definitive} bests even at this scope. This plurality of solid modeling options is encouraging, and suggests flexibility in the application of CDEKS to a range of use cases.

% \todo{ASW: one question this result raises for me is whether specific models do better at summarizing particular aspects of an event and might be preferred in contexts where those aspects are particularly important---e.g. time or location information, agent information, etc.}

% Broadly, we find raters' preferences to be variable, with clear winners failing to emerge even for a particular rater. For one rater, the only reliable preference ($p \leq 0.05$) was for the summaries of GPT-4o over those of T5. And a second rater reliably preferred the outputs of PEGASUS to those of BART, GPT-4o, and even the reference, but not to those of T5, GPT-4o Mini, or either Claude model.

%% file: sections/07-conclusion.tex
This work has extended the task of \emph{event-keyed summarization} (EKS) to the cross-document setting (CDEKS). To enable this, we provided an expert reannotation of the \famus CDAE dataset, yielding high-quality event argument annotations on all 1,265 examples. We then leveraged these improved annotations to construct \dataset---a collection of single- (\emph{report}) and cross-document summaries on top of \famus, further annotating the summaries themselves for event arguments (\S\ref{sec:annotation}). We benchmarked \dataset on a diverse set of baselines, including smaller fine-tuned models, as well as zero- and few-shot prompted LLMs (\S\ref{sec:experiments::report-summarization}, \S\ref{sec:experiments::cross-doc-summarization}). We then presented more detailed analysis, conducting a comprehensive set of input abalations (\S\ref{sec:experiments::input-ablations}), assessing the impact of degraded event extraction on summary quality (\S\ref{sec:experiments::extraction-quality}), and finally concluding with a human evaluation of summary quality (\S\ref{sec:analysis}). We release \dataset, along with our baseline results, to facilitate further work on EKS in both the single- and cross-document settings.

%% file: sections/08-limitations.tex
One limitation of this work is \dataset's size: 1,265 examples is sufficient for fine-tuning smaller models and for conducting prompting experiments with larger ones, but is likely insufficient for substantive fine-tuning of very large models.

% However, it was also precisely this restricted size that enabled us to obtain high-quality, in-house annotations for all components of the dataset.

A second limitation is that the cross-document setting considers only two documents per example. This constraint was imposed by the choice of the FAMuS dataset as the basis for \dataset, as cross-document argument annotations in the former were provided only for \emph{pairs} of report and source texts. Future work expanding the set of source texts would be valuable, and would allow both for richer summaries and for more robust evaluation of models' ability to accurately synthesize information across possibly differing accounts of events (cf.\ \citet{huang-etal-2024-embrace}), as information conflicts are more common outside of Wikipedia citations.

We note, however, that addressing either limitation may require relaxing data quality standards---relying on crowdsourcing or LLM-powered annotation techniques---as scaling our annotation procedure to many more examples or source documents would demand considerable resources. It was only thanks to the above restrictions that we were able to provide expert annotations for \dataset.

%% file: sections/09-ethics.tex
As the report and source texts in \dataset are the same as those in the FAMuS dataset, and as the summaries in \dataset are simply distillations of (parts of) these texts, we do not believe our dataset introduces any novel risks as a resource. Nonetheless, these texts do discuss real people, places, and institutions, and models trained on this data may thus be liable to make untrue claims about them or otherwise misrepresent them. We intend \dataset for academic use only, as a benchmark to evaluate systems for single- and cross-document event-keyed summarization.

%% file: appendices/additional_examples.tex
Below, we show a few examples of the report summaries and their corresponding cross-document summaries to illustrate how the latter typically provide greater detail about an event of interest relative to the former. We note, however, that this is not always the case: sometimes the source document offers no additional information about the event beyond what is contained in the report.

\paragraph{Example 1}
\begin{itemize}
    \item \textbf{Frame}: \textsc{Cause to Resume}
    \item \textbf{Report}: \emph{Areva \textbf{renewed} a uranium deal with Niger in January 2008.}
    \item \textbf{Cross-Doc}:  \emph{On January 13, 2008, French state-controlled nuclear reactor maker Areva CEPFi said it had \textbf{renewed} a uranium mining deal with the state of Niger and would invest over 1 billion euros.}
\end{itemize}

\paragraph{Exmaple 2}
\begin{itemize}
    \item \textbf{Frame}: \textsc{Smuggling}
    \item \textbf{Report}: \emph{A woman pled guilty to possession and attempting to \textbf{smuggle} 89 grams of heroin out of Thailand.}
    \item \textbf{Cross-Doc}: \emph{Scot Sandra Gregory pled guilty to possession and attempting to \textbf{smuggle} 89 grams of heroin out of Thailand in 1993 and did her time in Thai jails.}
\end{itemize}

\paragraph{Exmaple 3}
\begin{itemize}
    \item \textbf{Frame}: \textsc{Hostile Encounter}
    \item \textbf{Report}: \emph{The plot of Reign of Shadows involves players returning to the dark side of the moon of Luclin to \textbf{face} the snake-like Shissar race led by Emperor Ssraeshza.}
    \item \textbf{Cross-Doc}: \emph{The plot of Reign of Shadows involves players returning to the heart of the dark side of the moon of Luclin to \textbf{face} the snake-like Emperor Ssraeshza and his unyielding throngs of insidious zealots and enslaved minions to take back the ancient citadel of Vex Thal and end their march.}
\end{itemize}

%% file: appendices/training.tex
\paragraph{Models and Hardware} The BART, T5, and PEGASUS models were all trained on a single NVIDIA Quadro RTX 6000 GPU using CUDA version 11.7. Results reported with these models are based on single runs with a fixed random seed. We fine-tune the following pretrained checkpoints available from HuggingFace:
\begin{itemize}
    \item \texttt{t5-large}
    \setlength\itemsep{-0.5em}
    \item \texttt{facebook/bart-large}
    \item \texttt{google/pegasus-large}
\end{itemize}

\paragraph{Libraries} Models were developed using Python 3.11.9. We used the following libraries for model training, inference, and evaluation:
\begin{itemize}
    \setlength\itemsep{-0.5em}
    \item \texttt{accelerate (0.34.2)}
    \item \texttt{bert-score (0.3.13)}
    \item \texttt{bm25s (0.2.1)}
    \item \texttt{datasets (3.0.1)}
    \item \texttt{deepspeed (0.15.1)}
    \item \texttt{editdistance (0.8.1)}
    \item \texttt{evaluate (0.4.3)}
    \item \texttt{metametric (0.1.2)}
    \item \texttt{numpy (1.26.4)}
    \item \texttt{rouge-score (0.1.2)}
    \item \texttt{sentence-transformers (3.1.1)}
    \item \texttt{spacy (3.7.5)}
    \item \texttt{torch (2.0.1+cu117)}
    \item \texttt{transformers (4.45.1)}
    \item \texttt{tokenizers (0.20.0)}
\end{itemize}

\paragraph{Metrics} We use the implementations of ROUGE ($\textbf{R}_{1,2,L}$) and BERTScore (\textbf{BS}) provided by the HuggingFace \texttt{evaluate} library. We implement CEAF-REE (\textbf{CR}) and its soft-match variant (see Tables \ref{tab:report-only-ablation-results}, \ref{tab:combined-ablation-results}) using the \texttt{metametric} package \citep{chen-etal-2023-unified}. We use the implementation of AlignScore released by the metric's authors \citep{zha-etal-2023-alignscore}.\footnote{\url{https://github.com/yuh-zha/AlignScore}}. Lastly, for FActScore, we use the few-shot examples from \citet{wanner-etal-2024-closer} for decomposition and use Llama3.1-8B Instruct \citep{touvron2023llamaopenefficientfoundation, dubey2024llama3herdmodels} for both atomic fact decomposition and verification. 

\paragraph{Hyperparameters} BART, T5, and PEGASUS were all trained for a maximum of 30 epochs with a patience of 5 epochs, using ROUGE-1 ($\textbf{R}_1$) $\text{F}_1$ score on the dev set as the evaluation criterion. We use the Adam optimizer \citep{kingma-ba-2014-adam} with default hyperparameters ($\beta_1 = 0.9, \beta_2 = 0.99o$, $\epsilon=1e^{-8}$, $\eta=0.001$) for all models. For inference, we use beam search decoding with a beam size of 5 and set the maximum tokens to 256.

\paragraph{Input Formats}
Below, we show in greater detail the input format for BART, PEGASUS, and T5 for the report and cross-document results reported in \autoref{tab:merged-results} and \autoref{tab:corruption-merged}. (Note: these input formats were also used to obtain the BART, PEGASUS, and T5 results in the \textsc{Text+Event} rows in \autoref{tab:report-only-ablation-results} and \autoref{tab:combined-ablation-results}.) Here, \bos and \eos denote the model's start-of-sequence and end-of-sequence tokens, respectively (if applicable), and \sep denotes a special token used to delineate information pertaining to a particular event role. Other text set between angle brackets ($\langle \ldots \rangle$) denotes a variable placeholder. We add spaces between separators and adjacent text to improve readability below; they are not present in the actual input.

The input format for the \textbf{report} task is:

\begin{quote}
\bos Report: \ttset{Report Text} \eos \bos Frame \sep \ttset{Frame Name} \sep Trigger \sep \ttset{Trigger} \sep \ttset{Role 1 Name} \sep \ttset{Arg 1}; \ttset{Arg 2}; \ldots \sep \ttset{Role N Name} \sep \ttset{Arg 1}; \ttset{Arg 2}; \ldots \sep \eos
\end{quote}

The input format for the \textbf{cross-document} task is:

\begin{quote}
\bos Report: \ttset{Report Text} \sep Source: \ttset{Source Text} \eos \bos Report Event: Frame \sep \ttset{Frame Name} \sep Trigger \sep \ttset{Trigger} \sep \ttset{Role 1 Name} \sep \ttset{Arg 1}; \ttset{Arg 2}; \ldots \sep \ttset{Role N Name} \sep \ttset{Arg 1}; \ttset{Arg 2}; \ldots \sep Source Event: Frame \sep \ttset{Frame Name} \sep \ttset{Role 1 Name} \sep \ttset{Arg 1}; \ttset{Arg 2}; \ldots \sep \ttset{Role N Name} \sep \ttset{Arg 1}; \ttset{Arg 2}; \ldots \sep \eos
\end{quote}

The ablation settings presented in \autoref{tab:report-only-ablation-results} and \autoref{tab:combined-ablation-results} (\textsc{Text Only}, \textsc{Event Only}, \textsc{Text+Schema}) do not fundamentally change this overall structure, but merely omit parts of it (e.g.\ \textsc{Text+Schema} omits all \ttset{Arg N}).

%% file: appendices/llms.tex
\paragraph{GPT} All GPT models were accessed through the OpenAI Chat API\footnote{\url{https://platform.openai.com/docs/api-reference/chat}}, via the OpenAI Python SDK (\texttt{openai} 1.50.2). As noted in \S\ref{sec:experiments}, we set temperature to 0.7 and set the maximum output tokens to 256 (consistent with the fine-tuned models) for all experiments reported in this paper and leave the other API defaults unchanged ($n = 1$, top\_p is not set, and we use no frequency penalty, presence penalty, or logit bias). For GPT-4o, we used model version \texttt{gpt-4o-2024-08-06}. For GPT-4o Mini, we used model version \texttt{gpt-4o-mini-2024-07-18}. Results reported throughout the paper are based on a single generation per prompt.

\paragraph{Claude} All Claude models were accessed through the Anthropic Messages API\footnote{\url{https://docs.anthropic.com/en/api/messages}} via the Anthropic Python SDK (\texttt{anthropic} 0.34.2). As with the GPT models, we set temperature to 0.7 for all experiments in this paper and leave the other defaults unchanged (we do not set top\_p or top\_k, as recommended, and we do not set any stop sequences). For Claude 3.5 Sonnet, we used model version \texttt{claude-3-5-sonnet-20240620}. For Claude 3 Haiku, we used model version \texttt{claude-3-haiku-20240307}. Results reported throughout the paper are based on a single generation per prompt.

\paragraph{Prompts} We use the same prompts for all LLMs. Complete prompts are available in the public GitHub repository for this work. Here, we provide prompt templates used to obtain the results in \autoref{tab:merged-results} and \autoref{tab:corruption-merged}, for both tasks (report or cross-document) and for both the zero- (ZS) and few-shot (FS) settings. Text set between angle brackets $\langle \ldots \rangle$ denote placeholders.

We use the same system prompt for both tasks:

\begin{quote}
    You are an expert intelligence briefer. Your task is to analyze a specific, important event based ONLY on certain information, and to compile a concise summary of that event to be presented to a high-ranked decision maker.
\end{quote}

\noindent For the \textbf{report} task in the \textbf{zero-shot (ZS)} setting, the user prompt has the following structure:

\begin{quote}
The Report text below describes a situation. The Report Template provides specific details about the same situation. Focus ONLY on information relevant to the Situation Type.

Please write a short, accurate summary that is one sentence long and that is based ONLY on the provided information. DO NOT include any extraneous details. DO NOT use more than one sentence.

Situation Type: $\langle$\texttt{Frame Name}$\rangle$ ($\langle$\texttt{Frame Def}$\rangle$)

Report: $\langle$\texttt{Report Text}$\rangle$

Report Template:

- $\langle$\texttt{Role 1}$\rangle$ ($\langle$\texttt{Role 1 Def}$\rangle$): $\langle$\texttt{Arg 1}$\rangle$; $\langle$\texttt{Arg 2}$\rangle$; \ldots

- \ldots

- $\langle$\texttt{Role N}$\rangle$ ($\langle$\texttt{Role N Def}$\rangle$): $\langle$\texttt{Arg 1}$\rangle$; $\langle$\texttt{Arg 2}$\rangle$; \ldots

Summary:
\end{quote}
The \textbf{few-shot (FS)} user prompt for the \textbf{report} task had the following structure:
\begin{quote}
The Report text below describes a situation. The Report Template provides specific details about the same situation. Focus ONLY on information relevant to the Situation Type. 

Please write a short, accurate summary that is one sentence long and that is based ONLY on the provided information. DO NOT include any extraneous details. DO NOT use more than one sentence.

Here are a few examples to show you how to complete the task:

Example 1 \\
------------- \\
Situation Type: $\langle$\texttt{Frame Name}$\rangle$ ($\langle$\texttt{Frame Def}$\rangle$)

Report: $\langle$\texttt{Report Text}$\rangle$

Report Template:

- $\langle$\texttt{Role 1}$\rangle$ ($\langle$\texttt{Role 1 Def}$\rangle$): $\langle$\texttt{Arg 1}$\rangle$; $\langle$\texttt{Arg 2}$\rangle$; \ldots

- \ldots

- $\langle$\texttt{Role N}$\rangle$ ($\langle$\texttt{Role N Def}$\rangle$): $\langle$\texttt{Arg 1}$\rangle$; $\langle$\texttt{Arg 2}$\rangle$; \ldots

Summary: $\langle$\texttt{summary text}$\rangle$

Example 2 \\
------------- \\
$\langle$ \texttt{same format as above} $\rangle$

Example 3 \\
------------- \\
$\langle$ \texttt{same format as above} $\rangle$

Now here is the target example for you to complete:

Target \\
--------- \\
$\langle$\texttt{same format, but with summary text omitted}$\rangle$
\end{quote}

\noindent The \textbf{zero-shot} user prompt for the \textbf{cross-document} task had the following structure:
\begin{quote}
The Report text below describes a situation, and the Report Template provides specific details about the same situation. The Source text provides additional context about this situation, and the Source Template provides additional details. Focus ONLY on information relevant to the Situation Type.

Please write a short, accurate summary that is preferably one sentence long (and no more than two sentences long) based ONLY on the provided information. DO NOT include any extraneous details. TRY to use one sentence and DO NOT use more than two.

Situation Type: $\langle$\texttt{Frame Name}$\rangle$ ($\langle$\texttt{Frame Def}$\rangle$)

Report: $\langle$\texttt{Report Text}$\rangle$

Report Template:

- $\langle$\texttt{Role 1}$\rangle$ ($\langle$\texttt{Role 1 Def}$\rangle$): $\langle$\texttt{Arg 1}$\rangle$; $\langle$\texttt{Arg 2}$\rangle$; \ldots

- \ldots

- $\langle$\texttt{Role N}$\rangle$ ($\langle$\texttt{Role N Def}$\rangle$): $\langle$\texttt{Arg 1}$\rangle$; $\langle$\texttt{Arg 2}$\rangle$; \ldots

Situation Type: $\langle$\texttt{Frame Name}$\rangle$ ($\langle$\texttt{Frame Def}$\rangle$)

Source: $\langle$\texttt{Source Text}$\rangle$

Source Template:

- $\langle$\texttt{Role 1}$\rangle$ ($\langle$\texttt{Role 1 Def}$\rangle$): $\langle$\texttt{Arg 1}$\rangle$; $\langle$\texttt{Arg 2}$\rangle$; \ldots

- \ldots

- $\langle$\texttt{Role N}$\rangle$ ($\langle$\texttt{Role N Def}$\rangle$): $\langle$\texttt{Arg 1}$\rangle$; $\langle$\texttt{Arg 2}$\rangle$; \ldots

Summary:
\end{quote}

\noindent The \textbf{few-shot} user prompt for the \textbf{cross-document} task (not explicitly shown) follows exactly the same structure as the few-shot prompt for the report task, but naturally uses the cross-document example format in lieu of the report format.

%% file: appendices/human_eval.tex
Full instructions for the human evaluation, along with a JSON file containing the items that were rated, are provided in our GitHub repo (\url{https://github.com/wgantt/SEAMuS}).

%% file: appendices/annotation.tex
\subsection{License} We release \dataset and our code under a CC-BY-SA-4.0 license. As noted in the \textbf{Ethics} section, we intend \dataset for research use only, not for commercial purposes.

\subsection{Additional Summary Statistics} Additional summary statistics---about the report and source texts are shown in \autoref{tab:additional-summary-statistics}

\subsection{Inter-Annotator Agreement}
As we note in \S\ref{sec:annotation}, there was no redundancy in the \dataset annotation process: corrections to the FAMuS arguments, writing of summaries, and annotation of summary arguments were performed by a single annotator for each example. However, annotators did conduct a 10-example practice annotation for both the report and cross-document tasks. Thus, to give some (limited) sense of the inter-annotator agreement, \autoref{tab:report-summary-agreement} and \autoref{tab:combined-summary-agreement} present pairwise comparisons of annotators' annotations on these 10 items for the report and cross-document tasks (respectively) using the \emph{reference-based} metrics from \autoref{tab:merged-results} (plus the edit distance version of \textbf{CR}, \textbf{CR}\textsubscript{soft}; see \autoref{app:additional_results}). We treat annotations produced by annotators in the $P$ column as ``predictions'' to be evaluated against the ``reference'' annotations produced by annotators in the $R$ column. Two important notes:
\begin{enumerate}
    \item Because all of these metrics are $\text{F}_1$ scores, the distinction between $P$ and $R$ is moot and reversing $P$ and $R$ for any given pair would yield the same results. In both tables, we report results for all unordered annotator pairs, as well as the average across all pairs.
    \item Because these were practice annotations, none of them were included in the final \dataset dataset. We would thus expect the numbers reported here to be an \emph{underestimate} of the level of agreement on the main task, had we had redundancy.
\end{enumerate}

\begin{table}
    \centering
    \adjustbox{max width=\columnwidth}{
    \begin{tabular}{cc|cccccc}
    \toprule
        $P$ & $R$ & $\textbf{R}_1$ & $\textbf{R}_2$ & $\textbf{R}_L$ & \textbf{BS} & \textbf{CR} & $\textbf{CR}_\text{soft}$ \\
    \midrule
        A1 & A2 & $67.1$ & $44.5$ & $52.6$ & $93.4$ & $72.7$ & $86.9$ \\
        A1 & A3 & $72.4$ & $53.9$ & $63.1$ & $94.4$ & $76.4$ & $86.5$ \\
        A2 & A3 & $77.2$ & $60.5$ & $62.5$ & $94.7$ & $75.9$ & $89.6$\\
    \midrule
        \multicolumn{2}{c}{Avg.} & $72.2$ & $53.0$ & $59.2$ & $94.2$ & $75.0$ & $87.7$ \\
    \bottomrule
    \end{tabular}
    }
    \caption{Inter-annotator agreement on the 10 practice examples from the \dataset \textbf{report summary} annotation, as given by the reference-based metrics we report in \S\ref{sec:experiments}, treating annotator $P$'s responses as predictions and $R$'s responses as references (the reverse is equivalent, since these metrics are symmetric).}
    \label{tab:report-summary-agreement}
\end{table}

\begin{table}
    \centering
    \adjustbox{max width=\columnwidth}{
    \begin{tabular}{cc|cccccc}
    \toprule
        $P$ & $R$ & $\textbf{R}_1$ & $\textbf{R}_2$ & $\textbf{R}_L$ & \textbf{BS} & \textbf{CR} & $\textbf{CR}_\text{soft}$ \\
    \midrule
        A1 & A2 & $64.8$ & $42.4$ & $53.3$ & $94.0$ & $40.5$ & $56.6$ \\
        A1 & A3 & $64.7$ & $43.8$ & $51.6$ & $93.0$ & $50.6$ & $65.0$ \\
        A1 & A4 & $45.8$ & $22.7$ & $33.0$ & $90.7$ & $49.8$ & $65.2$ \\
        A1 & A5 & $69.0$ & $48.1$ & $58.8$ & $94.6$& $46.8$ & $62.8$ \\
        A2 & A3 & $77.7$ & $66.8$ & $72.4$ & $94.9$ & $50.6$ & $65.6$ \\
        A2 & A4 & $55.4$ & $33.9$ & $40.8$ & $91.2$ & $50.7$ & $66.8$ \\
        A2 & A5 & $72.0$ & $56.1$ & $64.9$ & $94.1$ & $50.6$ & $66.9$ \\
        A3 & A4 & $55.4$ & $33.2$ & $41.9$ & $90.7$ & $51.8$ & $69.0$ \\
        A3 & A5 & $71.0$ & $57.2$ & $61.8$ & $93.6$ & $52.6$ & $69.8$ \\
        A4 & A5 & $48.2$ & $27.2$ & $37.3$ & $90.7$ & $52.8$ & $69.6$ \\
        \midrule 
        \multicolumn{2}{c}{Avg.} & $62.4$ & $43.1$ & $51.6$ & $92.8$ & $49.7$ & $65.7$ \\
    \bottomrule
    \end{tabular}
    }
    \caption{Inter-annotator agreement on the 10 practice examples from the \dataset \textbf{cross-document summary} annotation, as given by the reference-based metrics we report in \S\ref{sec:experiments}, treating annotator $P$'s responses as predictions and $R$'s responses as references (the reverse is equivalent, since these metrics are symmetric).}
    \label{tab:combined-summary-agreement}
\end{table}

\begin{table}
    \adjustbox{max width=\columnwidth}{%
    \centering
    \begin{tabular}{l|rrrr}
    \toprule
         & \multicolumn{2}{c}{\textbf{Report}} & \multicolumn{2}{c}{\textbf{Source}} \\
    \cmidrule(lr){2-3} \cmidrule(lr){4-5}
         & Train & Dev & Train & Dev \\
    \midrule
        Examples & 759 & 253 & 759 & 253 \\
        Avg.\ Words & 59 & 60 & 1,084 & 1,511 \\
        Avg.\ Sentences & 2.0 & 2.0 & 44.7 & 61.5 \\
        Avg.\ Arguments & 3.1 & 3.5 & 3.8 & 4.2 \\
    \bottomrule
    \end{tabular}}
    \caption{Summary statistics for the \dataset report (left) and source documents, which are the same as those in the \famus dataset, albeit with slightly different arguments due to our corrections of the original \famus argument annotations.}
    \label{tab:additional-summary-statistics}
\end{table}

\subsection{Annotation Interface}
Here, we include screenshots of the annotation interface used to complete the Phase 2 annotation.\footnote{Recall that the Phase 1 annotation, which involved correcting the \famus report text argument annotations and writing the report summaries, was done in JSON files.} As noted in \S\ref{sec:annotation}, the interface was adapted from \citeauthor{vashishtha-etal-2024-famus}'s (\citeyear{vashishtha-etal-2024-famus}) annotation interface for the \famus cross-document argument extraction task (cf.\ Figures 5 and 6 in Appendix A of their paper). Tasks were run via Turkle, an open-source tool with similar functionality to Amazon Mechanical Turk.\footnote{\url{https://github.com/hltcoe/turkle-client}}

\begin{figure}
    \centering
    \includegraphics[width=\linewidth]{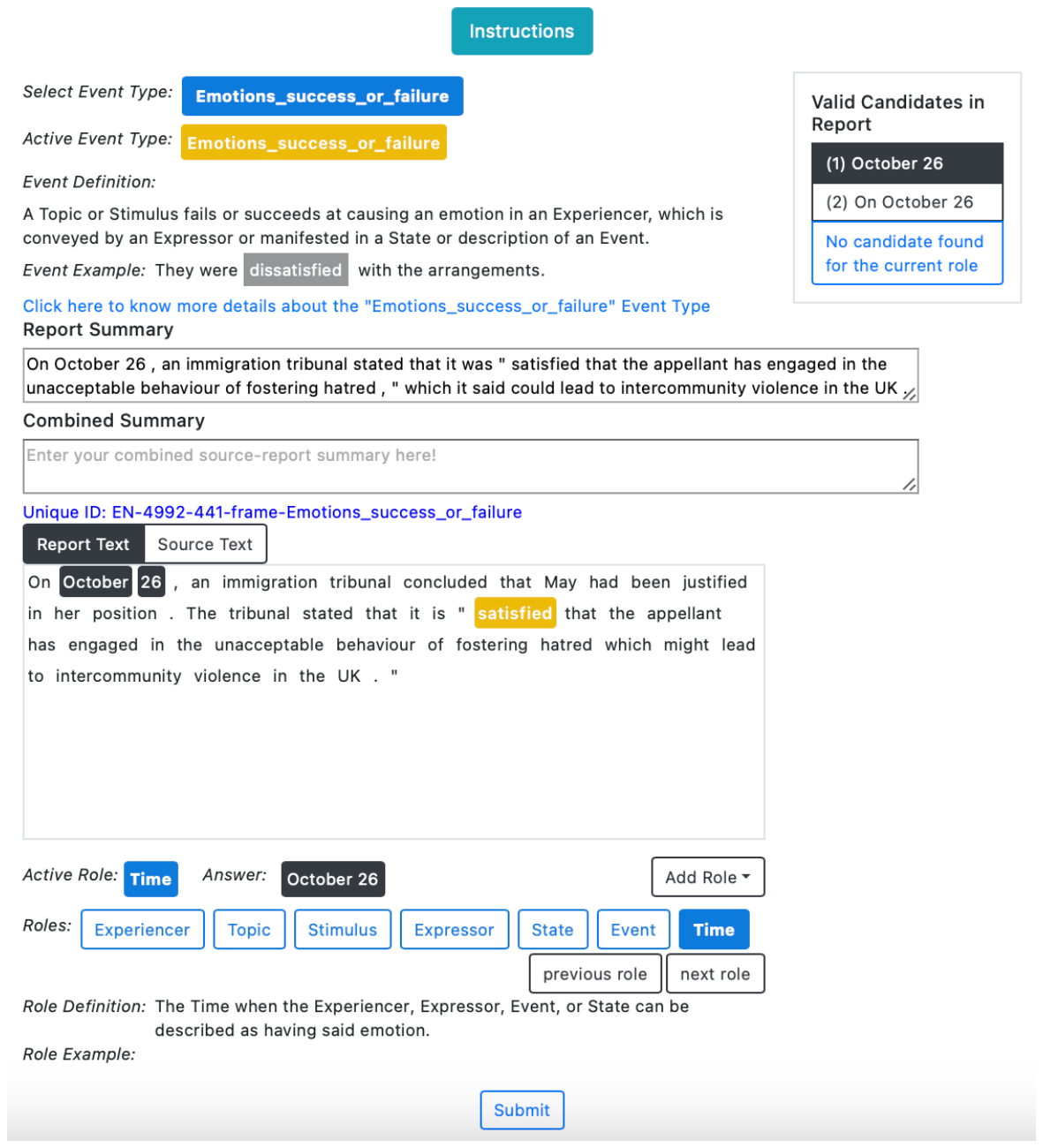}
    \caption{Interface for source text argument correction and cross-document summary writing (the first part of the Phase 2 annotation).\vspace{-8mm}}
    \label{fig:cross-doc-summary-writing}
\end{figure}

In the first part of the Phase 2 annotation, the existing (crowdsourced) \famus argument annotations for the source text were reviewed and corrected, and the cross-document summaries were written jointly on the basis of these corrected annotations and the corrected report text argument annotations from Phase 1 (see \autoref{fig:cross-doc-summary-writing}). The interface was pre-populated with (a) the corrected report text arguments from Phase 1 (in the ``Report Text'' tab, highlighted); the report summary from Phase 1 (in the ``Report Summary'' field); and (c) the uncorrected source text arguments (in the ``Source Text'' tab). The source text arguments were reviewed and corrected by toggling to the ``Source Text'' tab and making any necessary edits to the existing selections. The cross-document summaries were then written in the ``Combined Summary'' field. The UI for selecting, adding, and removing arguments was unchanged relative to \citeauthor{vashishtha-etal-2024-famus}'s implementation. The major differences here are the addition of the ``Report Summary'' and ``Combined Summary'' fields, and the inability to alter the selected FrameNet frame for annotation.

\begin{figure}
    \centering
    \includegraphics[angle=270, width=\linewidth]{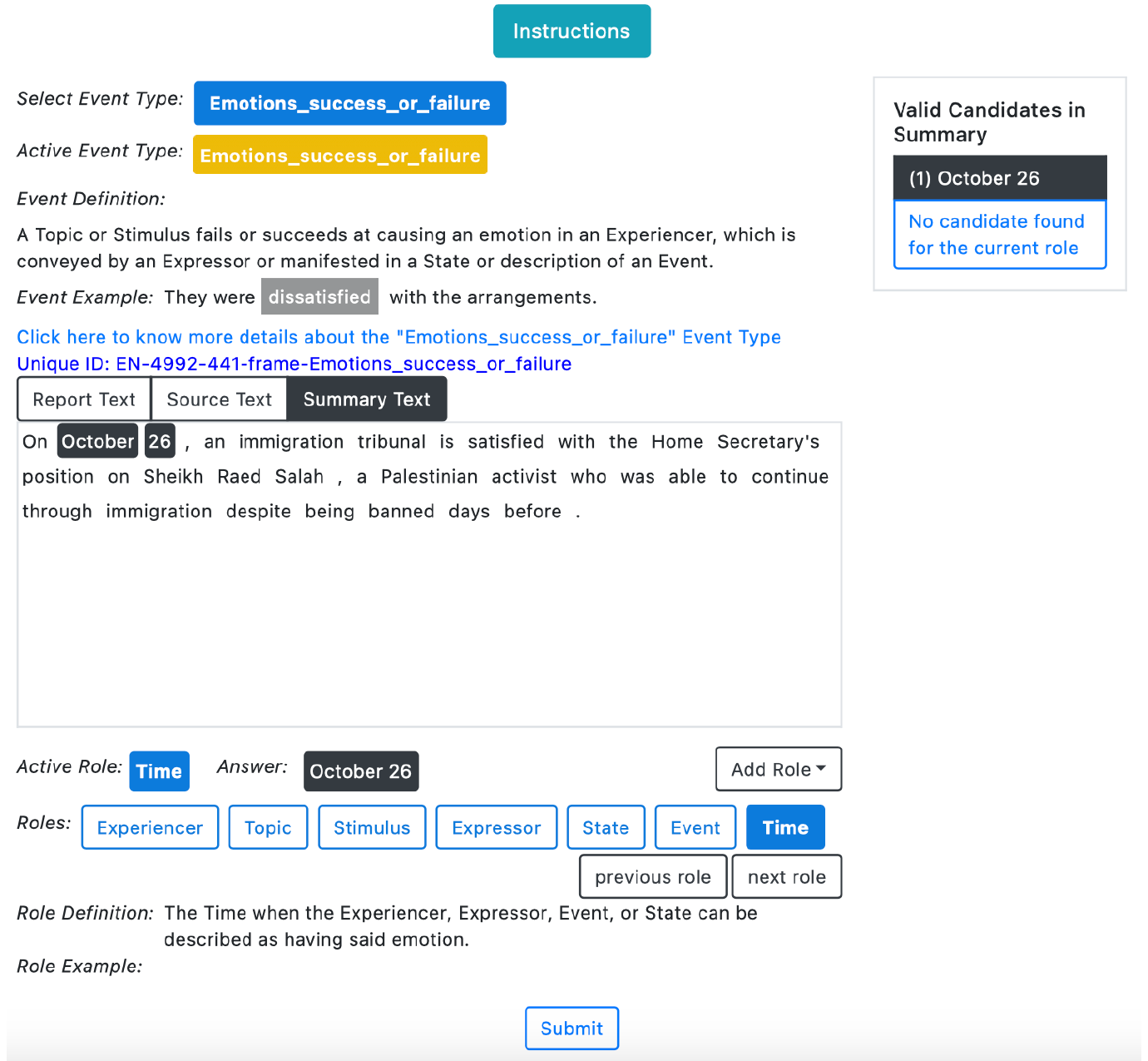}
    \caption{Interface for annotation of arguments on the cross-document summaries (the second part of the Phase 2 annotation).}
    \label{fig:cross-doc-arg-annotation}
\end{figure}

In the second part, arguments were annotated on the summaries written in the first part (\autoref{fig:cross-doc-arg-annotation}). The interface is similar to the interface for the first part of the Phase 2 annotation, except that the ``Report Summary'' and ``Combined Summary'' fields have been removed, and a new tab (``Summary Text'') containing the cross-document summary to be annotated was added. Summary arguments were annotated by toggling to this tab and making argument selections in the same way as before. Here, the corrected argument annotations for \emph{both} the report text \emph{and} for the source text were pre-populated for each task under their respective tabs, allowing annotators to toggle between these for reference in annotating the summary arguments.

As can be seen in both \autoref{fig:cross-doc-summary-writing} and \autoref{fig:cross-doc-arg-annotation}, details about the frame for the target event, including the frame name, its definition, as well as role names and their definitions, were provided as in the original \famus interface. Instructions were also accessible at any time via the dropdown shown at the top of the screen.

\subsection{Annotation Instructions}
Annotation instructions for both phases are available on our GitHub repo (\url{https://github.com/wgantt/SEAMuS}).

\subsection{Annotator Demographics}
The full set of annotators consists of six students (five graduate and one undergraduate) pursuing degrees in Computer Science (3), Linguistics (2), and Cognitive Science (1), all of whom are fluent English speakers. Only one was financially compensated for the annotations (at a rate of \$15 per hour), as this person initially became involved with the project through a university job board posting for the task, whereas the others were members of the lab from which the project originated. The project, and the intended use of their annotations, was clearly explained to all participants in meetings before they began any annotation.

%% file: appendices/additional_results.tex
\subsection{Main Results} \autoref{tab:report-cis} and \autoref{tab:cross-doc-cis} contain 95\% confidence intervals of the results in \autoref{tab:merged-results} based on non-parametric bootstraps ($n=1,000$).

\subsection{Input Ablations}
\label{app:additional-results::input-ablations}
Here, we include the full results of the ablations on the inputs introduced briefly in \S\ref{sec:experiments::input-ablations}, which were inspired by similar ones conducted by \citet{gantt-etal-2024-event}. In the \textsc{Text Only} setting, we omit information about the target event entirely and include only the text in the input---either the report for the report task, or both the report and source for the cross-document task---effectively reducing the problem to standard summarization. In the \textsc{Event Only} setting, we omit the text(s) and include only information about the target event---either the report event annotations for the report task, or both the report and source event annotations for the cross-document task---making this ablation similar to structure-to-text tasks, such as AMR-to-text \citep{pourdamghani-etal-2016-generating}). In the \textsc{Text+Schema} setting, we omit the argument annotations, but leave in information about the frame and its roles. For the fine-tuned models, we include just the names of the frame and its roles. For the LLMs, we additionally include the definitions of the frame and roles as given in FrameNet. Finally, \textsc{Text+Event} is the name we assign to the \emph{unablated} setting, used to obtain the results in \autoref{tab:merged-results} and \autoref{tab:corruption-merged}, where both the text(s) and the full event annotations are present in the input. For all ablation settings, BART, PEGASUS, and T5 are fine-tuned on the ablated inputs using the same settings for training and inference as are described in \S\ref{sec:experiments}. For the GPT and Claude models, the examples provided in the few-shot setting are also ablated in the way called for by each ablation.

\paragraph{Report} Results for the report task are in \autoref{tab:report-only-ablation-results}. Here and in the cross-document results to follow (\autoref{tab:combined-ablation-results}), we include a variant of CEAF-REE (\textbf{CR}) that we dub \textbf{CR}\textsubscript{soft}, which aligns and scores predicted arguments against reference arguments using normalized levenshtein distance rather than exact match---enabling a more nuanced comparison of different models' ability to recover event arguments in the summaries they produce.

Across all models and most metrics, we see significant drops in performance when ablating any component of the input. Notably, a number of models, especially the LLMs, fall to numbers near or below those of the report baseline (\textsc{RB}) on a variety of metrics.

There are, however, some unsurprising exceptions here. First, in many cases, results on \textbf{CR} and \textbf{CR}\textsubscript{soft} in the \textsc{Event Only} ablation are markedly stronger than the report baseline, and are even competitive with the results in the unablated setting (\textsc{Text+Event}) for most of the zero-shot-evaluated LLMs. This echoes a similar finding by \citet{gantt-etal-2024-event}, who note that ``the document [is not] needed to generate \emph{some} string that contains all the [event] template’s arguments.'' If this is correct, we would \emph{expect} to see strong \textbf{CR} scores in the \textsc{Event Only} setting, even though the summaries may be poorer overall (as reflected in other metrics).

An intriguing, related observation is that whereas the fine-tuned models look dominant against the LLMs on \textbf{CR} in the unablated setting, this advantage sharply diminishes when we turn to \textbf{CR}\textsubscript{soft}. This is likely explained by the fact that the fine-tuned models are able to learn the conventions adopted by annotators in selecting argument spans, whereas the (prompted) LLMs do not---even though they may still be generating outputs with approximately correct spans that are nonetheless harshly penalized by an exact match.

A second exception is the results on AlignScore (\textbf{A}) and FActScore (\textbf{F}) in the \textsc{Text Only} setting, which are competitive with---and in some cases superior to---the results in the unablated setting across models. Recall that both \textbf{A} and \textbf{F} here evaluate how well the report summary is supported by the report text. It is thus intuitively possible, and evidently quite feasible, to generate a summary that is adequately supported by the text without relying at all on the event annotations---which is exactly what is demanded by the \textsc{Text Only} setting. This is once again consistent with findings from \citet{gantt-etal-2024-event} on the NLI-based family of metrics MENLI \citep{chen-eger-2023-menli}, which are broadly similar to AlignScore and FActScore: ``[event] templates are not needed to generate \emph{some} summary that is entailed by the document.''

We also note that, for the fine-tuned models, we obtain \textbf{A} scores in the \textsc{Text+Schema} ablation that are comparable (T5) or higher than (BART, PEGASUS) those of the unablated setting. This makes sense, inasmuch as the \textsc{Text+Schema} setting contains a superset of the inputs of the \textsc{Text Only} setting, though it is unclear why we do not find a similar pattern with the LLMs.

Finally, note that the report baseline, which treats the report text itself as the summary, should in theory achieve perfect \textbf{A} and \textbf{F} scores, and thus does not really represent a fair comparison with the other models (note: this is also true for the cross-document setting). That it does not is surely a reflection of the fact that both metrics rely on outputs from imperfect models. Such flaws of LM-based metrics must not be overlooked. 

\paragraph{Cross-Document} results on the cross-document task are shown in \autoref{tab:combined-ablation-results} and follow a pattern that is qualitatively very similar to that of the report results above. We consistently find that the best results are obtained in the unablated setting (\textsc{Text+Event}) for most metrics, with the same exception regarding \textbf{CR}/\textbf{CR}\textsubscript{soft} in the \textsc{Event Only} setting as we found for the report task. Curiously, however, the findings on \textbf{A} are more complicated here: whereas we continue to see the strongest results on this metric in the \textsc{Text Only} and \textsc{Text+Schema} ablations for the fine-tuned models, with the LLMs, we instead see our best results in the unablated setting---following the trend of other metrics.

\begin{table*}
\centering
\adjustbox{max width=\textwidth}{
\begin{tabular}{lllllllllll}
\toprule
          \textbf{Model} & 
          \textbf{Ablation} &
          \textbf{Setting} & $\textbf{R}_1$ & $\textbf{R}_2$ & $\textbf{R}_L$ & \textbf{BS} & \textbf{CR} & 
          \textbf{CR}\textsubscript{soft} & 
          \textbf{A} & \textbf{F} \\
\midrule
% bart-1337: 83.34
% claude-3-5-sonnet-20240620-0-shot: 77.85
% claude-3-haiku-20240307-0-shot: 77.05
% gpt-4o-2024-08-06-0-shot: 73.05
% gpt-4o-mini-2024-07-18-0-shot: 76.38
% pegasus-1337: 84.33
% t5-1337: 82.93
Report Baseline & - & - & $56.15$ & $46.05$ & $48.37$ & $91.57$ & $52.58$ & $62.56$ & $99.11$ & $98.73$ \\
\midrule
\multirow{4}{*}{\textsc{GPT-4o M}} & \textonly & \zs & $49.96$ & $28.18$ & $39.23$ & $91.31$ & $34.59$ & $53.13$ & $95.74^\ast$ & $83.11$\\
& \eventonly & \zs & $53.11$ & $34.04$ & $43.67$ & $91.51$ & $52.13$ & $77.37$ & $60.98$ & $53.42$\\
& \textwithschema & \zs & $53.29$ & $31.60$ & $42.91$ & $91.28$ & $38.24$ & $56.92$ & $79.07$ & $76.38$\\
& \textwithevent & \zs & $62.18$ & $42.32$ & $51.26$ & $93.17$ & $58.48$ & $78.71$ & $86.04$ & $75.80$\\
& \textwithevent & \fs & $71.98$ & $55.35$ & $61.03$ & $94.34$ & $66.80$ & $83.66$ & $94.06$ & $83.32$\\
\midrule
\multirow{4}{*}{\textsc{GPT-4o}} & \textonly & \zs & $51.52$ & $29.90$ & $40.90$ & $91.50$ & $33.75$ & $52.06$ & $94.49$ & $84.00^\ast$\\
& \eventonly & \zs & $56.39$ & $38.34$ & $46.34$ & $91.93$ & $59.35$ & $83.35$ & $70.66$ & $57.14$\\

& \textwithschema & \zs & $56.57$ & $37.19$ & $47.08$ & $92.00$ & $42.37$ & $61.50$ & $81.66$ & $73.05$\\
& \textwithevent & \zs & $63.95$ & $45.21$ & $52.95$ & $93.18$ & $61.39^\ast$ & $82.60^\ast$ & $83.87$ & $74.78$\\
& \textwithevent & \fs & $72.54^\dag$ & $56.59^\dag$ & $62.34^\dag$ & $94.40$ & $69.61^\dag$ & $\mathbf{87.27}^\dag$ & $94.72$ & $81.58$\\
\midrule
\multirow{4}{*}{\textsc{Claude H}} & \textonly & \zs & $50.41$ & $30.39$ & $40.53$ & $91.11$ & $32.35$ & $51.46$ & $93.10$ & $83.77$ \\
& \eventonly & \zs & $55.03$ & $36.37$ & $45.71$ & $91.79$ & $54.36$ & $78.25$ & $72.15$ & $56.29$\\
& \textwithschema & \zs & $57.67$ & $38.51$ & $47.68$ & $92.08$ & $41.36$ & $59.10$ & $83.24$ & $77.05$\\
& \textwithevent & \zs & $64.75$ & $46.19$ & $54.67$ & $93.44$ & $58.75$ & $78.92$ & $84.87$ & $77.57$\\
& \textwithevent & \fs & $71.73$ & $55.86$ & $61.05$ & $94.29$ & $63.21$ & $80.95$ & $94.82$ & $82.54$\\
\midrule
\multirow{4}{*}{\textsc{Claude S}} & \textonly & \zs & $46.98$ & $22.83$ & $36.24$ & $90.78$ & $25.68$ & $45.88$ & $91.31$ & $82.41$\\
& \eventonly & \zs & $55.66$ & $36.89$ & $46.21$ & $92.13$ & $56.38$ & $78.54$ & $72.15$ & $60.37$\\
& \textwithschema & \zs & $57.33$ & $36.18$ & $46.98$ & $92.30$ & $41.71$ & $61.46$ & $88.93$ & $77.85$\\
& \textwithevent & \zs & $67.38^\ast$ & $48.11^\ast$ & $56.52^\ast$ & $93.84^\ast$ & $61.07$ & $81.35$ & $92.96$ & $80.59$\\
& \textwithevent & \fs & $72.16$ & $54.64$ & $61.29$ & $94.54^\dag$ & $65.66$ & $83.68$ & $95.89^\dag$ & $83.86^\dag$\\
\midrule
\multirow{4}{*}{\textsc{BART}} & \textonly & \ft & $57.13$ & $43.53$ & $50.46$ & $91.77$ & $46.27$ & $58.59$ & $97.42$ & $84.64$\\
& \eventonly & \ft & $58.34$ & $40.96$ & $48.51$ & $91.83$ & $59.82$ & $75.34$ & $51.17$ & $52.41$\\
& \textwithschema & \ft & $62.23$ & $49.43$ & $55.55$ & $92.59$ & $52.92$ & $65.83$ & $95.01$ & $83.34$\\
& \textwithevent & \ft & $74.46$ & $61.68$ & $66.42$ & $94.57$ & $69.88$ & $82.72$ & $91.59$ & $79.25$\\
\midrule
\multirow{4}{*}{\textsc{PEGASUS}} & \textonly & \ft & $60.33$ & $46.19$ & $52.44$ & $92.13$ & $45.95$ & $60.40$ & $97.45$ & $85.20$\\
& \eventonly & \ft & $59.69$ & $41.97$ & $49.46$ & $91.90$ & $57.14$ & $74.34$ & $53.93$ & $53.43$\\
& \textwithschema & \ft & $63.28$ & $49.79$ & $55.91$ & $92.71$ & $53.69$ & $66.28$ & $96.94$ & $84.33$\\
& \textwithevent & \ft & $75.18$ & $62.53$ & $66.96$ & $94.70$ & $70.00$ & $82.68$ & $96.08$ & $82.23$\\
\midrule
\multirow{4}{*}{\textsc{T5}} & \textonly & \ft & $58.38$ & $45.25$ & $51.81$ & $91.96$ & $49.70$ & $60.75$ & $\mathbf{98.88}$ & $\mathbf{87.85}$\\
& \eventonly & \ft & $63.14$ & $45.62$ & $52.47$ & $92.67$ & $64.00$ & $80.08$ & $68.42$ & $62.63$\\
& \textwithschema & \ft & $65.82$ & $51.90$ & $58.46$ & $93.11$ & $56.18$ & $68.42$ & $97.92$ & $82.93$\\
& \textwithevent & \ft & $\mathbf{76.64}$ & $\mathbf{64.44}$ & $\mathbf{68.90}$ & $\mathbf{95.02}$ & $\mathbf{74.20}$ & $85.22$ & $98.15$ & $85.02$\\
\bottomrule
\end{tabular}}
\caption{Input ablation results for the \textbf{report} summarization task. Best overall results are in \textbf{bolded}. $^\ast$ and $^\dag$ denote best zero- and few-shot results, respectively. See \S\ref{sec:experiments::overview} for an explanation of metrics. See \autoref{app:additional_results} for an explanation of the settings.}
\label{tab:report-only-ablation-results}
\end{table*}

%%%%%%%%%%%%%%%%%%%%
% COMBINED RESULTS %
%%%%%%%%%%%%%%%%%%%%
\begin{table*}
\centering
\adjustbox{max width=\textwidth}{
\begin{tabular}{lllllllllll}
\toprule
          \textbf{Model} & 
          \textbf{Ablation} &
          \textbf{Setting} & $\textbf{R}_1$ & $\textbf{R}_2$ & $\textbf{R}_L$ & \textbf{BS} & \textbf{CR} & 
          \textbf{CR}\textsubscript{soft} &
          \textbf{A} & \textbf{F}\\
\midrule
% ### VERIFY T5 ###
% ### VERIFY PEGASUS ###
% ### SCORING T5 & PEGASUS ###
% bart-bm25_concat_7-1337: 84.73
% claude-3-5-sonnet-20240620-0-shot: 91.46
% claude-3-5-sonnet-20240620-few-shot: 90.22
% claude-3-haiku-20240307-0-shot: 91.55
% claude-3-haiku-20240307-few-shot: 90.55
% gpt-4o-2024-08-06-0-shot: 88.72
% gpt-4o-2024-08-06-few-shot: 88.71
% gpt-4o-mini-2024-07-18-0-shot: 88.17
% gpt-4o-mini-2024-07-18-few-shot: 89.86
% t5-bm25_concat_7-1337: 90.19
% pegasus-bm25_concat_7-1337: 90.48

Report Baseline & - & - & $48.52$ & $33.28$ & $39.31$ & $89.58$ & $31.00$ & $42.04$ & $99.29$ & $93.12$\\
% Report Baseline (S) & - & $73.06$ & $63.19$ & $69.72$ & $94.47$ \\
\midrule
\multirow{4}{*}{\textsc{GPT-4o M}} & \textonly & \zs & $37.56$ & $16.93$ & $26.97$ & $88.98$ & $21.86$ & $40.48$ & $73.58$ & $91.60$\\
& \eventonly & \zs & $52.45$ & $31.15$ & $40.04$ & $91.17$ & $37.48$ & $66.51$ & $69.97$ & $75.00$\\
& \textwithschema & \zs & $41.88$ & $20.40$ & $30.32$ & $89.72$ & $24.04$ & $44.76$ & $76.64$ & $89.12$\\
& \textwithevent & \zs & $51.87$ & $29.90$ & $39.10$ & $91.31$ & $38.99$ & $64.13$ & $81.46$ & $88.89$\\
& \textwithevent & \fs & $57.48$ & $36.99$ & $45.74$ & $92.08$ & $39.78$ & $62.93$ & $88.48$ & $89.79$\\
\midrule
\multirow{4}{*}{\textsc{GPT-4o}} & \textonly & \zs & $41.59$ & $19.28$ & $30.70$ & $89.48$ & $21.60$ & $42.04$ & $69.09$ & $92.06$\\
& \eventonly & \zs & $54.03$ & $33.98$ & $42.13$ & $91.51$ & $41.75^\ast$ & $\mathbf{69.63}^\ast$ & $81.02$ & $80.55$\\
& \textwithschema & \zs & $49.87$ & $27.04$ & $37.76$ & $90.86$ & $25.80$ & $48.53$ & $85.44$ & $89.75$\\
& \textwithevent & \zs & $57.97$ & $36.42$ & $45.89$ & $92.22^\ast$ & $41.34$ & $68.04$ & $86.61$ & $88.41$\\
& \textwithevent & \fs & $61.17^\dag$ & $40.62^\dag$ & $49.38^\dag$ & $\mathbf{92.67}^\dag$ & $42.72^\dag$ & $69.27^\dag$ & $90.62$ & $88.45$\\
\midrule
\multirow{4}{*}{\textsc{Claude H}} & \textonly & \zs & $47.27$ & $25.48$ & $36.49$ & $90.23$ & $22.64$ & $43.20$ & $84.29$ & $92.59$\\
& \eventonly & \zs & $53.35$ & $33.01$ & $42.94$ & $91.39$ & $38.64$ & $66.08$ & $77.70$ & $76.83$\\
& \textwithschema & \zs & $51.79$ & $30.45$ & $41.04$ & $90.87$ & $26.38$ & $48.02$ & $87.10$ & $90.87$\\
& \textwithevent & \zs & $57.72^\ast$ & $36.88^\ast$ & $46.35^\ast$ & $92.05$ & $36.22$ & $60.03$ & $90.37$ & $91.36$\\
& \textwithevent & \fs & $59.42$ & $39.40$ & $48.56$ & $92.13$ & $37.20$ & $59.70$ & $90.99$ & $90.50^\dag$\\
\midrule
\multirow{4}{*}{\textsc{Claude S}} & \textonly & \zs & $44.13$ & $20.08$ & $32.73$ & $89.88$ & $19.93$ & $40.24$ & $87.26$ & $92.30$\\
& \eventonly & \zs & $53.51$ & $33.51$ & $42.73$ & $91.53$ & $39.78$ & $66.17$ & $84.12$ & $81.91$\\
& \textwithschema & \zs & $51.37$ & $29.33$ & $40.06$ & $90.94$ & $28.05$ & $49.07$ & $88.64$ & $89.33$\\
& \textwithevent & \zs & $56.77$ & $34.75$ & $45.27$ & $91.91$ & $35.24$ & $59.47$ & $93.41^\ast$ & $91.71^\ast$\\
& \textwithevent & \fs & $57.95$ & $38.05$ & $47.53$ & $92.09$ & $37.32$ & $59.31$ & $95.09^\dag$ & $90.39$\\
\midrule
\multirow{4}{*}{\textsc{BART}} & \textonly & \ft & $48.57$ & $30.30$ & $39.70$ & $89.99$ & $27.12$ & $44.43$ & $90.06$ & $86.87$\\
& \eventonly & \ft & $56.37$ & $37.04$ & $45.14$ & $91.21$ & $39.12$ & $62.90$ & $56.01$ & $68.10$\\
& \textwithschema & \ft & $51.67$ & $35.12$ & $44.15$ & $90.42$ & $32.31$ & $49.47$ & $94.45$ & $90.52$\\
& \textwithevent & \ft & $63.77$ & $45.50$ & $52.98$ & $92.59$ & $44.97$ & $66.36$ & $85.55$ & $85.27$\\
\midrule
\multirow{4}{*}{\textsc{PEGASUS}} & \textonly & \ft & $50.85$ & $33.44$ & $42.51$ & $90.29$ & $30.22$ & $47.46$ & $97.63$ & $91.80$\\
& \eventonly & \ft & $58.52$ & $38.41$ & $46.46$ & $91.42$ & $39.98$ & $64.06$ & $67.05$ & $75.80$\\
& \textwithschema & \ft & $51.21$ & $34.18$ & $43.11$ & $90.28$ & $30.15$ & $47.04$ & $97.99$ & $\mathbf{92.72}$\\
& \textwithevent & \ft & $63.66$ & $46.24$ & $\mathbf{53.18}$ & $92.51$ & $43.73$ & $64.51$ & $93.85$ & $90.48$\\
\midrule
\multirow{4}{*}{\textsc{T5}} & \textonly & \ft & $49.18$ & $33.15$ & $41.39$ & $89.94$ & $30.98$ & $46.58$ & $\mathbf{98.75}$ & $91.60$\\
& \eventonly & \ft & $59.96$ & $40.55$ & $47.51$ & $91.84$ & $\mathbf{45.30}$ & $68.85$ & $73.73$ & $78.98$\\
& \textwithschema & \ft & $53.06$ & $35.64$ & $44.93$ & $90.64$ & $31.87$ & $50.14$ & $94.11$ & $91.30$\\
& \textwithevent & \ft & $\mathbf{64.14}$ & $\mathbf{46.36}$ & $52.79$ & $92.56$ & $44.67$ & $65.66$ & $92.48$ & $90.19$\\
\bottomrule
\end{tabular}}
\caption{Input ablations on the \textbf{cross-document} summarization task.Best overall results are in \textbf{bolded}. $^\ast$ and $^\dag$ denote best zero- and few-shot results, respectively. See \S\ref{sec:experiments::overview} for an explanation of metrics. See \autoref{app:additional_results} for an explanation of the settings.}
\label{tab:combined-ablation-results}
\end{table*}

\begin{table*}[t]
\centering
\adjustbox{max width=\textwidth}{
\begin{tabular}{ll|ccccccc}
\toprule
& & \multicolumn{7}{c}{\textbf{Report}}\\
\cmidrule(lr){1-9}
\textbf{Model} & \textbf{S} & $\textbf{R}_1$ & $\textbf{R}_2$ & $\textbf{R}_L$ & \textbf{BS} & \textbf{CR} & \textbf{A} & \textbf{F}\\
\midrule
\textsc{GPT-4o M} & \zs & \conf{62.2}{60.0}{64.5} & \conf{42.2}{39.3}{45.2} & \conf{51.2}{48.7}{54.0} & \conf{93.2}{92.8}{93.6} & \conf{56.7}{52.8}{60.5} & \conf{83.8}{80.5}{86.9} & \conf{75.8}{72.8}{78.6}\\
& \fs & \conf{71.9}{69.8}{74.0} & \conf{55.3}{52.5}{58.1} & \conf{61.0}{58.5}{63.5} & \conf{94.3}{94.0}{94.7} & \conf{64.9}{60.7}{68.8} & \conf{94.7}{93.4}{95.8} & \conf{83.3}{80.8}{85.8} \\
% \midrule
\textsc{GPT-4o} &
\zs & \conf{63.9}{61.6}{66.3} & \conf{45.1}{42.4}{48.2} & \conf{52.9}{50.5}{55.5} & \conf{93.2}{92.8}{93.6} & \conf{58.4}{54.2}{62.3} & \conf{86.0}{83.0}{88.8} & \conf{74.7}{71.2}{78.0}\\
& \fs & \conf{72.5}{70.3}{74.9} & \conf{56.6}{53.6}{59.8} & \conf{62.3}{59.6}{65.0} & \conf{94.4}{94.0}{94.8} & \conf{66.8}{62.7}{70.5} & \conf{94.0}{92.6}{95.4} & \conf{81.5}{78.4}{84.5}\\
% \midrule
\textsc{Claude H} & \zs & \conf{64.8}{62.7}{67.2} & \conf{46.1}{43.5}{49.1} & \conf{54.7}{52.4}{57.3} & \conf{93.4}{93.1}{93.9} & \conf{56.4}{52.5}{60.4} & \conf{84.8}{81.5}{87.7} & \conf{77.5}{74.1}{80.1} \\
& \fs & \conf{71.7}{69.4}{73.9} & \conf{55.7}{52.7}{58.7} & \conf{61.0}{58.5}{63.6} & \conf{94.3}{93.9}{94.7} & \conf{62.3}{58.1}{66.5} & \conf{94.8}{93.3}{96.1} & \conf{82.5}{79.6}{85.2}\\
% \midrule
\textsc{Claude S} & \zs & \conf{67.3}{65.1}{69.6} & \conf{48.0}{45.3}{50.8} & \conf{56.5}{54.1}{59.0} & \conf{93.8}{93.5}{94.2} & \conf{59.2}{55.0}{63.3} & \conf{92.9}{90.8}{94.8} & \conf{80.6}{77.6}{83.5}\\
& \fs & \conf{72.0}{69.8}{74.4} & \conf{54.5}{51.7}{57.5} & \conf{61.3}{58.8}{53.8} & \conf{94.5}{94.2}{94.9} & \conf{64.6}{60.7}{68.6} & \conf{95.9}{94.8}{96.7} & \conf{83.9}{80.8}{86.5} \\
% \midrule
\textsc{BART} & FT & \conf{74.3}{71.9}{76.6} & \conf{61.7}{58.7}{64.6} & \conf{66.4}{63.7}{69.1} & \conf{93.7}{93.3}{94.1} & \conf{68.2}{64.3}{72.1} & \conf{91.6}{89.2}{93.9} & \conf{79.2}{76.1}{82.2}\\
% \midrule
\textsc{PEGASUS} & FT & \conf{75.1}{72.9}{77.5} & \conf{62.4}{59.5}{65.4} & \conf{66.8}{64.2}{69.5} & \conf{93.7}{93.3}{94.1} & \conf{69.0}{65.4}{72.4} & \conf{96.1}{94.4}{97.5} & \conf{82.2}{79.4}{85.0}\\
% \midrule
\textsc{T5} & FT & \conf{76.6}{74.3}{78.9} & \conf{64.3}{61.4}{67.3} & \conf{68.8}{66.1}{71.5} & \conf{94.0}{93.6}{94.4} & \conf{73.3}{69.7}{76.9} & \conf{98.2}{97.4}{98.8} & \conf{85.0}{82.4}{87.5}\\
\bottomrule
\end{tabular}}
\caption{95\% confidence intervals \conf{\text{mean}}{\text{low}}{\text{high}} from a non-parametric bootstrap ($n=1000$) of the \textbf{report} results given in \autoref{tab:merged-results}.}
\label{tab:report-cis}
\end{table*}

\begin{table*}
\centering
\adjustbox{max width=\textwidth}{
\begin{tabular}{ll|ccccccc}
\toprule
& & \multicolumn{7}{c}{\textbf{Cross-Document}} \\
\cmidrule(lr){1-9}
\textbf{Model} & \textbf{S} & $\textbf{R}_1$ & $\textbf{R}_2$ & $\textbf{R}_L$ & \textbf{BS} & \textbf{CR} & \textbf{A} & \textbf{F} \\
\midrule
\textsc{GPT-4o M} & \zs &  \conf{51.8}{49.9}{53.7} & \conf{29.8}{27.8}{32.0} & \conf{39.0}{37.0}{41.0} & \conf{91.3}{91.0}{91.6} & \conf{39.2}{35.9}{42.2} & \conf{81.4}{78.5}{84.1} & \conf{88.8}{86.7}{90.7}\\
& \fs & \conf{47.4}{55.2}{59.7} & \conf{36.8}{34.3}{39.5} & \conf{45.7}{43.4}{47.8} & \conf{92.1}{91.7}{92.4} & \conf{39.4}{35.6}{42.9} & \conf{88.4}{86.0}{90.6} & \conf{89.7}{87.7}{91.6} \\
% \midrule
\textsc{GPT-4o} &
\zs & \conf{57.9}{55.7}{60.0} & \conf{36.2}{33.8}{38.9} & \conf{45.7}{43.5}{48.1} & \conf{92.2}{91.8}{92.6} & \conf{39.9}{36.4}{43.5} & \conf{86.5}{83.8}{89.0} & \conf{88.4}{86.0}{90.6}\\
& \fs & \conf{61.1}{59.0}{63.3} & \conf{40.6}{38.1}{43.1} & \conf{49.3}{47.1}{51.5} & \conf{92.7}{92.3}{93.0} & \conf{41.4}{38.0}{45.3} & \conf{90.6}{88.4}{92.8} & \conf{88.4}{86.3}{90.5}\\
% \midrule
\textsc{Claude H} & \zs &  \conf{57.7}{55.6}{59.7} & \conf{36.8}{34.3}{39.2} & \conf{46.4}{44.0}{48.7} & \conf{92.0}{91.7}{92.4} & \conf{36.3}{32.8}{39.9} & \conf{90.4}{88.1}{92.3} & \conf{91.3}{89.7}{92.9}\\
& \fs & \conf{59.4}{57.0}{61.5} & \conf{39.4}{36.7}{42.1} & \conf{48.5}{46.0}{50.9} & \conf{92.1}{91.8}{92.5} & \conf{36.8}{33.4}{40.4} & \conf{91.0}{89.0}{92.9} & \conf{90.5}{88.8}{92.2}\\
% \midrule
\textsc{Claude S} & \zs & \conf{56.7}{54.7}{58.9} & \conf{34.7}{32.4}{37.2} & \conf{45.3}{43.1}{47.7} & \conf{91.9}{91.6}{92.3} & \conf{34.5}{31.1}{37.9} & \conf{93.4}{91.7}{95.0} & \conf{91.7}{89.6}{93.4}\\
& \fs & \conf{57.9}{55.6}{60.4} & \conf{38.0}{35.4}{40.8} & \conf{47.5}{45.2}{49.9} & \conf{92.1}{91.7}{92.5} & \conf{37.6}{33.9}{41.5} & \conf{95.1}{94.1}{96.0} & \conf{90.4}{88.5}{92.2}\\
% \midrule
\textsc{BART} & FT & \conf{63.8}{61.5}{66.1} & \conf{45.5}{42.7}{48.4} & \conf{53.0}{50.5}{55.6} & \conf{91.9}{91.5}{92.2} & \conf{45.1}{41.3}{49.1} & \conf{85.6}{82.3}{88.6} & \conf{85.3}{82.6}{87.7}\\
% \midrule
\textsc{PEGASUS} & FT & \conf{63.6}{61.2}{66.0} & \conf{46.0}{43.1}{49.0} & \conf{53.1}{50.4}{55.8} & \conf{91.7}{91.3}{92.1} & \conf{44.7}{40.9}{48.4} & \conf{93.8}{91.7}{95.6} & \conf{90.5}{88.7}{92.3}\\
% \midrule
\textsc{T5} & FT & \conf{64.0}{61.5}{66.4} & \conf{46.2}{43.6}{49.2} & \conf{52.8}{50.2}{55.3} & \conf{91.8}{91.3}{92.2} & \conf{44.5}{40.3}{48.4} & \conf{92.4}{90.1}{94.4} & \conf{90.2}{88.2}{91.9}\\
\bottomrule
\end{tabular}}
\caption{95\% confidence intervals \conf{\text{mean}}{\text{low}}{\text{high}} from a non-parametric bootstrap ($n=1000$) of the \textbf{cross-document} results given in \autoref{tab:merged-results}.}
\label{tab:cross-doc-cis}
\end{table*}

\subsection{Argument Recovery by Role}
\autoref{tab:cross-doc-arg-recovery-hard} and \autoref{tab:cross-doc-arg-recovery-soft} show \textbf{CR} and \textbf{CR}\textsubscript{soft} results (respectively) on the cross-document task broken down by role for the 20 roles with highest support (number of annotated arguments) in the \dataset training split.

Comparing the tables reveals an interesting dichotomy. For \textbf{CR}, no model is consistently dominant across all roles, with fine-tuned models collectively obtaining the best results on 12 of the 20 and few-shot prompted models obtaining the best results on the remaining 8. The \textbf{CR}\textsubscript{soft} results, by contrast, heavily favor \textsc{GPT-4o}, which achieves the best scores on 13 roles. Here, the fine-tuned models are top-performing on only 4 roles.

We believe the same factor discussed in \autoref{app:additional-results::input-ablations} explains this dichotomy: whereas \textbf{CR} requires exact span match---and thus will tend to favor models able to learn span boundary conventions through fine-tuning---\textbf{CR}\textsubscript{soft} does not, and rewards spans proportional to their edit distance from the reference. Thus, \textbf{CR}\textsubscript{soft} reveals the LLMs (and \textsc{GPT-4o} above all) to be effective in producing summaries that recover the correct arguments, albeit with more lexical modifications relative to the reference.

\begin{table*}
    \centering
    \adjustbox{max width=\textwidth}{
    \begin{tabular}{lllllllll}
    \toprule
        \textbf{Role} & \textbf{Support} & \textbf{\textsc{GPT-4o M}} & \textbf{\textsc{GPT-4o}} & \textbf{\textsc{Claude H}} & \textbf{\textsc{Claude S}} & \textbf{\textsc{BART}} & \textbf{\textsc{PEGASUS}} & \textbf{\textsc{T5}} \\
    \midrule
    \textsc{Time} & $523$ & $39.13$ & $40.69$ & $34.17$ & $37.25$ & $42.07$ & $44.32$ & $\mathbf{47.09}$\\
    \textsc{Place} & $499$ & $33.33$ & $38.49$ & $25.00$ & $26.12$ & $27.42$ & $33.11$ & $\mathbf{38.56}$\\
    \textsc{Agent} & $240$ & $34.67$ & $32.89$ & $27.40$ & $23.13$ & $\mathbf{42.38}$ & $32.43$ & $39.74$\\
    \textsc{Theme} & $94$ & $\mathbf{49.12}$ & $44.07$ & $43.33$ & $35.09$ & $40.00$ & $39.44$ & $39.34$\\
    \textsc{Entity} & $65$ & $29.27$ & $35.90$ & $35.00$ & $35.90$ & $30.00$ & $35.90$ & $\mathbf{41.03}$\\
    \textsc{Patient} & $53$ & $41.18$ & $30.30$ & $50.00$ & $34.29$ & $43.75$ & $\mathbf{51.61}$ & $48.48$ \\
    \textsc{Goal} & $49$ & $36.84$ & $\mathbf{50.00}$ & $37.84$ & $27.03$ & $30.00$ & $40.91$ & $45.00$ \\
    \textsc{Event} & $43$ & $14.81$ & $20.69$ & $20.69$ & $\phantom{0}7.14$ & $18.75$ & $\mathbf{25.00}$ & $\phantom{0}6.45$ \\
    \textsc{Cause} & $42$ & $\phantom{0}6.45$ & $24.24$ & $11.43$ & $12.12$ & $31.25$ & $10.81$ & $\mathbf{26.67}$ \\
    \textsc{Experiencer} & $39$ & $38.10$ & $\mathbf{70.00}$ & $\mathbf{70.00}$ & $54.55$ & $52.17$ & $38.46$ & $54.55$ \\
    \textsc{Victim} & $39$ & $41.38$ & $\mathbf{53.33}$ & $48.28$ & $34.48$ & $31.25$ & $37.50$ & $32.26$\\
    \textsc{Goods} & $38$ & $26.67$ & $\mathbf{75.00}$ & $\phantom{0}0.00$ & $28.57$ & $50.00$ & $50.00$ & $14.29$\\
    \textsc{Protagonist} & $38$ & $26.67$ & $37.50$ & $37.50$ & $40.00$ & $40.00$ & $37.50$ & $\mathbf{50.00}$  \\
    \textsc{Source} & $30$ & $66.67$ & $\mathbf{77.78}$ & $55.56$ & $63.16$ & $63.16$ & $50.00$ & $66.67$\\
    \textsc{Topic} & $26$ & $13.33$ & $13.33$ & $\phantom{0}0.00$ & $14.29$ & $\mathbf{15.38}$ & $13.33$ & $\phantom{0}0.00$ \\ 
    \textsc{Speaker} & $25$ & $50.00$ & $50.00$ & $66.67$ & $50.00$ & $50.00$ & $\mathbf{70.59}$ & $58.82$\\
    \textsc{Addressee} & $22$ & $33.33$ & $\mathbf{60.00}$ & $16.67$ & $40.00$ & $\mathbf{60.00}$ & $40.00$ & $33.33$ \\
    \textsc{Stimulus} & $21$ & $15.38$ & $30.77$ & $33.33$ & $33.33$ & $46.15$ & $33.33$ & $\mathbf{50.00}$\\
    \bottomrule
    \end{tabular}}
    \caption{\textbf{CR} $\text{F}_1$ results on test set \textbf{cross-document} summaries for the top 20 roles with highest support (\# arguments) in the \dataset training split (which has 3,004 total arguments). Results with GPT and Claude models are from the few-shot (\textsc{FS}) setting. Best results for each role are \textbf{bolded}.}
    \label{tab:cross-doc-arg-recovery-hard}
\end{table*}

\begin{table*}
    \centering
    \adjustbox{max width=\textwidth}{
    \begin{tabular}{lllllllll}
    \toprule
        \textbf{Role} & \textbf{Support} & \textbf{\textsc{GPT-4o M}} & \textbf{\textsc{GPT-4o}} & \textbf{\textsc{Claude H}} & \textbf{\textsc{Claude S}} & \textbf{\textsc{BART}} & \textbf{\textsc{PEGASUS}} & \textbf{\textsc{T5}} \\
    \midrule
    \textsc{Time} & $523$ & $57.37$ & $65.22$ & $49.22$ & $50.52$ & $60.54$ & $61.40$ & $\mathbf{65.43}$\\
    \textsc{Place} & $499$ & $44.84$ & $\mathbf{53.39}$ & $37.46$ & $37.67$ & $46.93$ & $47.67$ & $52.15$\\
    \textsc{Agent} & $240$ & $65.40$ & $\mathbf{67.47}$ & $59.78$ & $50.16$ & $66.72$ & $57.99$ & $62.06$\\
    \textsc{Theme} & $94$ & $73.21$ & $\mathbf{73.96}$ & $69.22$ & $72.96$ & $71.75$ & $64.71$ & $68.59$\\
    \textsc{Entity} & $65$ & $66.33$ & $\mathbf{76.13}$ & $63.16$ & $60.07$ & $63.34$ & $63.12$ & $69.44$\\
    \textsc{Patient} & $53$ & $\mathbf{76.28}$ & $73.52$ & $74.41$ & $68.62$ & $70.88$ & $73.28$ & $73.36$ \\
    \textsc{Goal} & $49$ & $50.53$ & $\mathbf{66.30}$ & $50.19$ & $45.42$ & $47.30$ & $53.63$ & $60.30$ \\
    \textsc{Event} & $43$ & $52.75$ & $\mathbf{63.52}$ & $52.76$ & $42.95$ & $45.97$ & $52.09$ & $35.38$ \\
    \textsc{Cause} & $42$ & $47.66$ & $\mathbf{52.99}$ & $39.41$ & $42.03$ & $43.23$ & $43.87$ & $49.82$ \\
    \textsc{Experiencer} & $39$ & $68.83$ & $\mathbf{91.48}$ & $82.86$ & $69.44$ & $56.03$ & $60.99$ & $61.06$ \\
    \textsc{Victim} & $39$ & $62.13$ & $\mathbf{74.79}$ & $71.75$ & $60.13$ & $65.62$ & $60.53$ & $55.89$\\
    \textsc{Goods} & $38$ & $42.77$ & $\mathbf{79.33}$ & $33.35$ & $50.68$ & $61.94$ & $57.46$ & $24.39$\\
    \textsc{Protagonist} & $38$ & $60.13$ & $\mathbf{72.52}$ & $54.02$ & $66.03$ & $66.57$ & $67.63$ & $61.24$  \\
    \textsc{Source} & $30$ & $78.00$ & $\mathbf{84.13}$ & $60.54$ & $65.55$ & $66.80$ & $55.03$ & $69.62$\\
    \textsc{Topic} & $26$ & $19.06$ & $19.06$ & $17.75$ & $20.42$ & $21.70$ & $\mathbf{35.80}$ & $12.96$ \\ 
    \textsc{Speaker} & $25$ & $58.21$ & $61.41$ & $71.85$ & $64.91$ & $64.92$ & $\mathbf{73.03}$ & $66.28$\\
    \textsc{Addressee} & $22$ & $44.60$ & $\mathbf{89.49}$ & $40.21$ & $52.63$ & $63.63$ & $73.95$ & $41.94$ \\
    \textsc{Stimulus} & $21$ & $38.61$ & $70.46$ & $54.55$ & $66.70$ & $53.39$ & $\mathbf{74.75}$ & $57.65$\\
    \bottomrule
    \end{tabular}}
    \caption{$\textbf{CR}_\text{soft}$ (distinct from \textbf{CR}; see \S\ref{app:additional-results::input-ablations}) $\text{F}_1$ results on test set \textbf{cross-document} summaries for the top 20 roles with highest support (\# arguments) in the \dataset training split (which has 3,004 total arguments). Results with the GPT and Claude models are from the few-shot (\textsc{FS}) setting. Best results for each role are \textbf{bolded}.}
    \label{tab:cross-doc-arg-recovery-soft}
\end{table*}

%% file: appendices/ai_assistants.tex
GitHub Copilot was used as a coding assistant for parts of model development and data analysis, though its suggestions were carefully reviewed by the authors. AI assistants were \textbf{not} used for other parts of this work (writing, brainstorming, etc.).